%% file: main.tex
\documentclass{article}

 \usepackage[eandd, final]{neurips_2026}

\usepackage[utf8]{inputenc} 
\usepackage[T1]{fontenc}    
\usepackage{hyperref}       
\usepackage{url}            
\usepackage{booktabs}       
\usepackage{amsfonts}       
\usepackage{nicefrac}       
\usepackage{microtype}      
\usepackage{xcolor}         

\input{preamble}

\RequirePackage{xspace}
\makeatletter
\DeclareRobustCommand\onedot{\futurelet\@let@token\@onedot}
\def\@onedot{\ifx\@let@token.\else.\null\fi\xspace}

\def\ie{\emph{i.e}\onedot}

\makeatother

\title{VLAN: Vision-Language Accessible Navigation}

\author{
  Jie Hu \\
  Hunan University\\
  \And
  Yu Zheng \\
  Hunan University\\
  \And
  Wenzhi Wu \\
  Hunan University\\
  \And
  Weixing Feng \\
  Hunan University\\
  \And
  Cunlei Cheng \\
  Hunan University\\
  \And
  Jiaming Zhang\thanks{Corresponding author, (\href{mailto:jiamingzhang@hnu.edu.cn}{jiamingzhang@hnu.edu.cn}).} \\
  Hunan University\\
}

\begin{document}

\maketitle

\begin{figure}[H]
\centering
\includegraphics[width=1.0\textwidth]{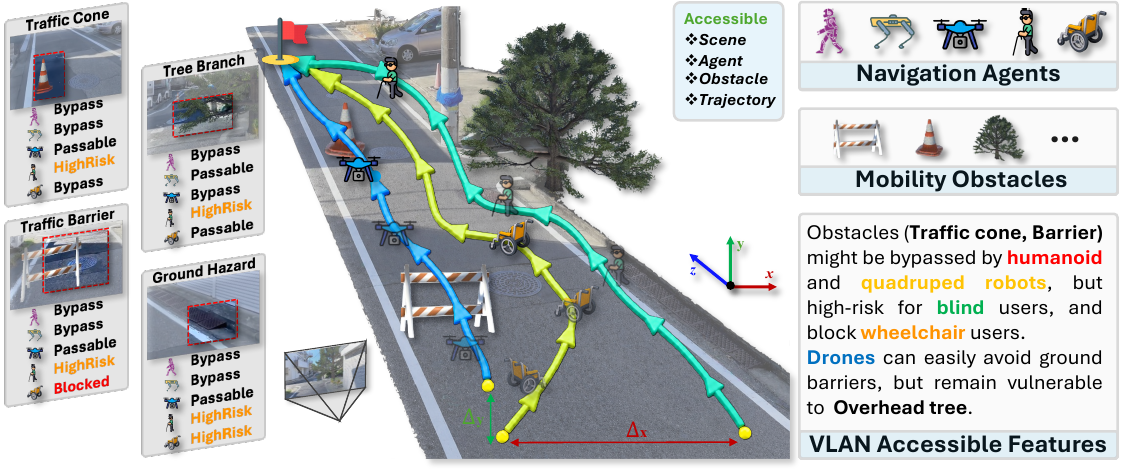}
\caption{\textbf{Vision-Language Accessible Navigation (VLAN)} evaluates whether models can produce embodiment-specific navigation actions and accessibility predictions for a visual scene with same start-goal pair and obstacle layout. Since accessibility is conditioned on user mobility and perception, the same obstacle may induce distinct outcomes across users, ranging from passable to blocked.}
\label{fig:fig1}
\end{figure}

\begin{abstract}
Accessible mobility is user-conditioned rather than scene-intrinsic, as walkable paths for one embodiment are inaccessible to another. Yet, existing benchmarks evaluate traversability under a fixed and limited embodiment. We present \textbf{Vision-Language Accessible Navigation (VLAN)}, containing \textbf{1000 scenes}, \textbf{20K navigation trajectories}, and \textbf{160K keyframes} across diverse lighting, environmental conditions, scene semantics, and path geometries. For each trajectory, VLAN provides embodiment-specific actions and accessibility features for legged and wheeled agents, including humanoid and quadruped robots and people with visual or mobility impairments, while everyday obstacles and accessibility labels are categorized as \textit{Passable}, \textit{Bypass}, \textit{HighRisk}, and \textit{Blocked}. Unlike goal-only evaluation, our benchmark measures physical safety and alignment with user-specific constraints via goal-reaching, collision-aware, and accessibility-aligned metrics. We benchmark modern vision-language models and find they often fail to consistently adapt obstacle relevance, path selection, and safety criteria across user profiles, suggesting accessibility-aware navigation requires dedicated datasets, tasks, and metrics beyond conventional embodied navigation evaluation, with all data and code open-source at:\url{https://github.com/InSAI-Lab/VLAN}.
\end{abstract}

\input{Sections/1_introduction}
\input{Sections/2_related_work}
\input{Sections/3_benchmark}

\input{Sections/4_experiments}

\input{Sections/5_conclusion}

\clearpage
\noindent \textbf{Acknowledgment} This work was supported in part by the National Natural Science Foundation of China under Grant No. 62503166, in part by the Hunan Provincial Research and Development Project under Grant number 2026QK3018, and in part by the Project of Yuelushan Center for Industrial Innovation. 
{
\small
\bibliographystyle{unsrt}
\bibliography{main}
}

\clearpage

\input{appendix}


\newpage

\end{document}

%% file: preamble.tex
\usepackage[dvipsnames, table]{xcolor}

\usepackage{algorithm}
\usepackage{algpseudocode}

\usepackage{graphicx}
\usepackage{amsmath}
\usepackage{amssymb}
\usepackage{booktabs}
\usepackage{mathtools}
\usepackage[accsupp]{axessibility}  
\usepackage{color}
\usepackage{array}
\usepackage{pdflscape}
\usepackage[longtable]{multirow}
\usepackage{longtable}
\usepackage{tabularray}
\usepackage{booktabs}
\usepackage{times}
\usepackage{epsfig}
\usepackage{wrapfig}
\usepackage{amsmath}
\usepackage{bbm}
\DeclareMathAlphabet\mathbfcal{OMS}{cmsy}{b}{n}
\usepackage{makecell}
\usepackage{siunitx}
\usepackage{float}
\usepackage{comment}
\usepackage{stfloats}
\usepackage{multicol}
\usepackage{multirow}
\usepackage{bm}
\usepackage{etoolbox}
\usepackage{icomma}
\usepackage{array}
\usepackage{tabulary}
\usepackage{xcolor}
\usepackage{paralist}
\usepackage{booktabs}
\usepackage{adjustbox}
\usepackage{caption}
\usepackage{subcaption}
\usepackage{arydshln}
\usepackage{rotating}
\usepackage{enumitem}

\definecolor{gray}{rgb}{0.3,0.3,0.3}
\definecolor{blue}{rgb}{0,0.5,1}
\definecolor{mask_red}{rgb}{1,0,0.8}
\definecolor{green}{rgb}{0.2,1,0.2}
\definecolor{rblue}{rgb}{0,0,1}
\definecolor{lightblue}{HTML}{6495ed}
\definecolor{lightred}{HTML}{F19C99}

\definecolor{graytablerow}{gray}{0.6}

\usepackage{pifont}
\usepackage{tabu}
\usepackage[pro]{fontawesome5}

\usepackage{tikz}

\usepackage[capitalize]{cleveref}
\crefname{section}{Sec.}{Secs.}
\Crefname{section}{Section}{Sections}
\Crefname{table}{Table}{Tables}
\crefname{table}{Tab.}{Tabs.}

%% file: Sections/1_introduction.tex
\section{Introduction} \label{intro}

Embodied navigation~\cite{song2025towards, ziliotto2025tango, zheng2024towards, gu2022vision} in real-world environments is governed not only by scene geometry and goal location, but also by the physical, perceptual, and operational constraints of the target user or agent. The same obstacle can induce substantially different navigation decisions across embodiments. As illustrated in Fig.~\ref{fig:fig1}, \textit{a low temporary barrier can be traversable for a humanoid robot, risky for a BVI user, and impassable for a user with mobility impairments}. Conversely, \textit{overhanging tree branches may have little effect on quadruped robots, yet become critical hazards for drones}. These cases highlight a limitation of user-agnostic navigation evaluation, since reaching a target is insufficient if the resulting trajectory is unsafe, inaccessible, or misaligned with the intended user.

Existing navigation benchmarks~\cite{ku2020room,krantz_vlnce_2020} have driven substantial progress in vision-language navigation, embodied navigation, and action prediction. However, they typically assume a fixed embodiment and a user-independent traversability criterion, where a trajectory is considered successful if it reaches the target and avoids collisions for a generic agent. This formulation is insufficient for accessibility-aware navigation~\cite{jiang2025bench}, where obstacle relevance, safety margins, action granularity, and feasible strategies depend on the target embodiment. Although recent capability-conditioned and embodiment-specific studies~\cite{su2026capnav, windecker2025navitrace, cheng2024navila, wang2024towards} have started to examine these differences, existing benchmarks still lack a unified setting for evaluating multiple human and robotic users in the same scenes with user-specific actions and accessibility annotations.

Building such a benchmark requires both realistic observations and interactive, editable environments. Real-world videos and street-level imagery capture natural scene appearance, but they are passive and do not readily support scene editing, counterfactual environment modification, action rollout, or collision checking. Simulation provides interactivity and control, but scaling visually diverse and realistic scenes remains costly. Recent work~\cite{xie2025vid2sim, han2025roomtour3d} converts monocular videos into photorealistic and physically interactable 3D simulation environments for embodied navigation, enabling reinforcement learning of visual navigation agents in digital twins. Inspired by these works, 
we find that editable reconstructed scenes can preserve realistic visual appearance while supporting interactive scene modification, trajectory generation, keyframe rendering, and metric-grounded evaluation. 

Built on these, we present \textbf{Vision-Language Accessible Navigation (VLAN)}, a large-scale dataset for evaluating multi-user accessibility-aware navigation. VLAN contains \textbf{1,000 scenes}, \textbf{20K navigation trajectories}, and \textbf{160K keyframes}, spanning diverse lighting, environmental conditions, scene semantics, and path geometries. 
We provide user-conditioned navigation actions and accessibility annotations for multiple user embodiments, \ie, \textit{humanoid robots}, \textit{quadruped robots}, \textit{drones}, \textit{blind or visually impaired (BVI) users}, and \textit{wheelchair users}. This design enables us to generate user-conditioned navigation instances from each scene without relying solely on additional real-world data collection. The drone setting is deliberately formulated as \emph{fixed-altitude 2D accessibility-conditioned navigation}: the action space is $(\Delta x, \Delta z, \Delta\theta)$ and overhead obstacles are handled by a horizontal detour or a stop, so that the drone and the quadruped robot share one action interface and remain directly comparable on the same scene.

Given the same visual path with start-goal pair, VLAN evaluates whether a model can reason and forecast future navigation actions from visual observations and action history for different user embodiments. We propose user-dependent accessibility-aligned metrics, enabling systematic analysis of whether modern vision-language models can adapt obstacle relevance, path selection, safety margins, and action behavior across different users. We instantiate the benchmark with general-purpose vision-language models rather than task-specific navigation agents: existing VLN agents assume a single fixed embodiment and consume an interactive egocentric observation stream, whereas VLAN provides a fixed set of keyframes and requires one model to switch its output modality and safety criteria across five user types. We therefore treat generic VLMs as the appropriate first baseline and expect embodiment-specialised navigation agents to be a natural next step.

Our contributions are summarized as follows:

\begin{itemize}

    \item We introduce \textbf{VLAN}, a large-scale benchmark for multi-user accessibility-aware navigation with 1,000 scenes, 20K trajectories, and 160K keyframes. For the first time, VLAN involves five embodiments and accessible features for evaluating inclusive navigation. 
    \item We formulate a user-conditioned navigation setting where the same scene and obstacle layout can induce different navigation actions and accessibility labels for different user embodiments, \ie, \textit{humanoid robots}, \textit{quadruped robots}, \textit{drones}, \textit{blind or visually impaired (BVI) users}, and \textit{wheelchair users}.
    \item We provide user-specific actions, obstacle categories, accessibility labels, and collision-aware evaluation protocols for assessing large vision-language models beyond goal-reaching performance.
    \item We benchmark modern vision-language models and show that current systems struggle to consistently adapt navigation behavior and safety criteria across user profiles.

\end{itemize}

%% file: Sections/2_related_work.tex
\section{Related Work} \label{related}
\noindent\textbf{Embodied and capability-conditioned navigation.}
Vision-language navigation benchmarks such as R2R~\cite{anderson2018vision}, RxR~\cite{ku2020room}, and VLN-CE~\cite{krantz_vlnce_2020} have advanced instruction-following navigation from discrete graph settings to continuous embodied environments. Recent work explores long-horizon VLN and scalable task generation~\cite{song2025towards}. However, most benchmarks assume a fixed embodiment and a user-independent notion of traversability. Capability- and embodiment-conditioned studies begin to address this limitation. CapNav~\cite{su2026capnav} evaluates VLMs under agent-specific mobility constraints, NaviTrace~\cite{windecker2025navitrace} studies embodiment-conditioned image-space navigation traces, and VLN-PE~\cite{wang2025rethinking} analyzes the physical and visual gaps in cross-embodiment VLN. Platform-specific work on legged robots and UAVs further shows that different agents require different action spaces and obstacle constraints~\cite{cheng2024navila,wang2024towards}. EmbodiedCity~\cite{gao2024embodiedcity} and Embodied4C~\cite{sohn2025embodied4c} further introduce embodiment shifts across vehicles, drones, and manipulators, but remain robot- or vehicle-centric: they include no human users with perceptual or mobility impairments, no user-conditioned accessibility labels, and no test of whether a model produces \emph{different} outputs for different users given identical visual input. VLAN differs by evaluating multiple human and robotic users in the same scenes with user-specific actions, accessibility labels, and safety-aware metrics.

\noindent\textbf{Assistive and accessibility-aware benchmarks.} Recent assistive benchmarks evaluate VLMs for visually impaired assistance and human-centered perception. VIALM shows that large models can lack environment grounding and fine-grained guidance for visually impaired users~\cite{zhao2024vialm}. @BENCH~\cite{jiang2025bench} evaluates VLMs on assistive tasks selected by people with visual impairments, including segmentation, depth estimation, OCR, captioning, and VQA. WalkVLM~\cite{yuan2025walkvlm} constructs a video-based benchmark for visually impaired walking assistance. These works emphasize accessibility-aware reasoning, but they are largely perception-, description-, or reminder-oriented. VLAN instead provides an action-level and instruction-level navigation benchmark with goal-reaching, collision-aware, and accessibility-aligned evaluation.

\noindent\textbf{Real-to-interactive simulation for navigation.}
Videos are visually diverse but passive, while simulation enables interaction but is costly to scale with realistic scenes. Established embodied platforms~\cite{savva2019habitat} provide control and photorealistic rendering but still require hand-built 3D assets. RoomTour3D~\cite{han2025roomtour3d} constructs geometry-aware video-instruction data from web-based room tour videos, while Vid2Sim~\cite{xie2025vid2sim} converts monocular videos into photorealistic and physically interactable simulation environments for urban navigation. EmbodiedSplat~\cite{chhablani2025embodiedsplat}, GaussGym~\cite{escontrela2025gaussgym}, and RoboGSim~\cite{li2024robogsim} further explore Gaussian-splatting-based interactive environments for robot learning and evaluation. VLAN builds on this direction for a different purpose, using editable reconstructed scenes to construct a scalable multi-user accessibility-aware navigation benchmark.

%% file: Sections/3_benchmark.tex
\section{VLAN Dataset and Benchmark}

\label{sec:dataset}
VLAN is a large-scale benchmark for multi-user accessibility-aware navigation. It contains 1,000 scenes, 20K navigation trajectories, and 160K keyframes. Each scene contains about 20 trajectories, and each trajectory is represented by eight keyframes. For each trajectory, VLAN provides visual observations, camera poses, obstacle metadata, accessibility labels, and user-conditioned navigation targets. The benchmark is designed to evaluate whether models can adapt navigation behavior to different user embodiments under the same visual scene, start-goal pair, and obstacle layout.

\subsection{Dataset Construction and Diversity} \label{subsec:dataset_construction}

\begin{figure}[H]
    \centering
    \includegraphics[width=1.0\textwidth]{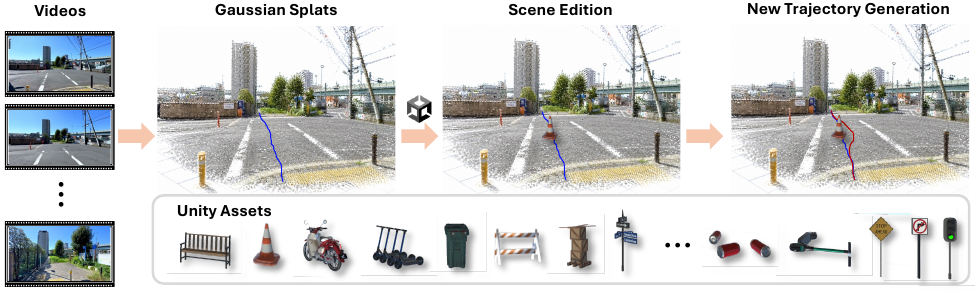}
    \caption{
    \textbf{VLAN scene reconstruction pipeline.} Egocentric videos are reconstructed into Gaussian scenes and converted into Unity. The resulting scenes support arbitrary editing, trajectory generation, keyframe rendering, and construction of user-conditioned navigation action or instructions.}
    \label{fig:data_pipe}
\end{figure}

\begin{wrapfigure}{r}{0.63\linewidth}
    \vspace{-1.0em}
    \centering
    \begin{subfigure}{0.36\linewidth}
        \centering
        \includegraphics[width=\linewidth]{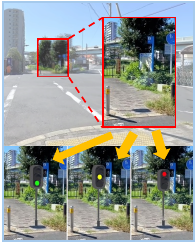}
        \caption{Traffic Light.}
        \label{fig:scene_edit-a}
    \end{subfigure}
    \hfill
    \begin{subfigure}{0.60\linewidth}
        \centering
        \includegraphics[width=\linewidth]{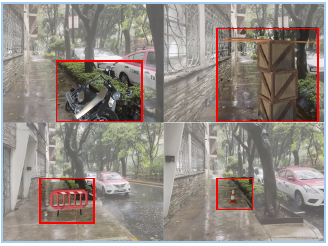}
        \caption{Navigation Obstacles.}
        \label{fig:scene_edit-b}
    \end{subfigure}

    \caption{\textbf{Editable scenes of VLAN.} The same reconstructed scene can be modified with arbitrary traffic-light states, obstacle layouts, and route conditions. It can systematically construct accessibility-aware navigation scenarios.} 
    \label{fig:scene_edit}
    \vspace{-1.0em}
\end{wrapfigure}

\noindent\textbf{Scene reconstruction and editing.}
VLAN is constructed from egocentric scene videos and associated navigation routes. Fig.~\ref{fig:data_pipe} summarizes the overall pipeline from video collection to Gaussian reconstruction~\cite{li2025streamgs, wang2024vggsfm, xie2025vid2sim}, Unity-based scene editing, trajectory generation, and sample construction. We first reconstruct each scene into a Gaussian-splat representation and import the reconstructed scene into Unity~\cite{juliani2018unity} for controlled editing. Instead of treating captured environments as fixed visual observations, VLAN uses reconstructed scenes as interactive and editable environments. This design preserves realistic visual appearance while enabling systematic variation of scene states, route conditions, and obstacle configurations. As shown in Fig.~\ref{fig:scene_edit}, the same reconstructed scene can be assigned different traffic-light states and modified with different obstacle configurations. This editability is important for data scaling because one base environment can produce multiple navigation instances with different accessibility implications. VLAN therefore scales along two complementary axes. Increasing the number of reconstructed scenes improves scene-level diversity across appearance, layout, lighting, and environmental conditions. Increasing the number of edited routes within each scene improves navigation-level diversity under shared visual geometry.

\noindent\textbf{Scene and obstacle diversity.}
VLAN covers diverse navigation conditions across lighting, scene type, path geometry, weather, and surface condition. The scene set includes daytime, nighttime, dusk, and low-vision settings, as well as urban, suburban, rural, and indoor environments. It also contains wide, open, narrow, and constrained routes under sunny, rainy, snowy, and wet-ground conditions. Table~\ref{tab:scene_taxonomy_summary} summarizes the scene taxonomy distribution over the 1,000 classified scenes. This diversity is important because accessibility depends on both the obstacle itself and the surrounding context. The same object may be easy to bypass in an open area but may block a narrow passage, while wet ground or poor visibility may further change route safety for users with visual or mobility constraints.

VLAN groups obstacles according to their functional role in navigation. \textbf{Navigation obstacles} include objects that directly affect route selection, such as barricades, cones, and temporary fences. \textbf{Ground obstacles} include low barriers, curbs, boxes, uneven surfaces, and other objects that influence ground mobility. \textbf{Overhead obstacles} include tree branches, hanging signs, and canopies. \textbf{Traffic-sign} consist of sign boards, poles, and roadside signs that can act as both semantic cues and physical obstacles. These categories reflect that obstacle relevance changes across user embodiments.

\input{Tables/0_scene_taxonomy}

\subsection{User-Conditioned Trajectories and Actions}
\label{subsec:user_conditioned_trajectories}
VLAN considers multiple human and robotic user embodiments, including humanoid robots, quadruped robots, drones, blind or visually impaired users, and wheelchair users. These users differ in body geometry, mobility, perception, safety requirements, and navigation interface. Humanoid and quadruped robots both move on the ground, but differ in height and body profile. Drones operate in aerial space and are mainly constrained by overhead hazards. BVI users require risk-aware guidance because some obstacles may be physically avoidable but difficult to perceive. Wheelchair users require physically accessible routes with sufficient clearance, stable ground, and no blocking ground-level barriers.

\noindent\textbf{Accessibility labels.}
After constructing the editable scenes, we manually annotate accessibility for each scene trajectory with respect to each target user embodiment. Each route is assigned one of four accessibility labels, \textit{Passable}, \textit{Bypass}, \textit{HighRisk}, and \textit{Blocked}. \textit{Passable} indicates that the route can be followed without special avoidance. \textit{Bypass} indicates that the user can proceed by taking a detour around the obstacle. \textit{HighRisk} indicates that the route may be physically possible but unsafe or unreliable for the target user. \textit{Blocked} indicates that the route is inaccessible and should not be followed. These labels are user-conditioned by design. For example, a low barrier may be bypassable for humanoid and quadruped robots, high-risk for BVI users, and blocked for wheelchair users, while an overhead branch is irrelevant to ground users but obstruct humanoid users, BVI users, and drones.

\noindent\textbf{Navigation routes generation.} Let $\mathcal{S}$ denote an edited scene, $o$ the obstacle configuration, $x_s$ the start state, $x_g$ the goal state, and $u$ the user type. We define a user-conditioned route as \begin{equation} \tau^{u} = \{x_1^{u}, x_2^{u}, \dots, x_T^{u}\} = \mathcal{G}(\mathcal{S}, o, x_s, x_g, u), \end{equation} where $\mathcal{G}$ denotes the trajectory generation process in Unity under user-specific accessibility constraints. In practice, $\mathcal{G}$ produces different route patterns depending on the target label. For \textit{Passable} routes, the generated path remains close to the nominal route; for \textit{Bypass} routes, the path explicitly detours around the obstacle; for \textit{HighRisk} routes, the route is generated more conservatively when a feasible corridor exists; and for \textit{Blocked} routes, the target behavior is to stop or reroute rather than force unsafe goal reaching. This design yields scene-consistent but user-specific trajectory supervision. The same edited environment can therefore produce multiple navigation routes and multiple accessibility interpretations under different user embodiments.

\noindent\textbf{Expert Action Construction.} 
For action-prediction users, we derive expert actions from the generated route. Each trajectory contains a sequence of poses
\begin{equation}
x_t = (p_t, \psi_t),
\end{equation}
where $p_t \in \mathbb{R}^3$ is the world position and $\psi_t$ is the yaw angle. We convert consecutive poses into local navigation actions of the form
\begin{equation}
a_t = (\Delta x_t, \Delta z_t, \Delta \theta_t),
\end{equation}
where $\Delta x_t$ is lateral displacement, $\Delta z_t$ is forward displacement, and $\Delta \theta_t$ is yaw rotation.

Let $R(\psi_t)$ be the planar rotation matrix defined by the current yaw. The action target is computed as
\begin{equation}
\begin{bmatrix}
\Delta x_t \\
\Delta z_t
\end{bmatrix}
=
R(\psi_t)^{\top}
\left(
\begin{bmatrix}
p_{t+1}^{x} \\
p_{t+1}^{z}
\end{bmatrix}
-
\begin{bmatrix}
p_t^{x} \\
p_t^{z}
\end{bmatrix}
\right),
\qquad
\Delta \theta_t = \psi_{t+1} - \psi_t.
\end{equation}
Vertical motion is not used in the present ground-plane action representation: the drone setting is therefore formulated as \emph{fixed-altitude 2D accessibility-conditioned navigation}, and its labels, reference actions, and metrics are all defined on the ground plane. We detail this design decision and its scope in Appendix~\ref{app:drone2d}.

The resulting expert actions encode accessibility-aware behavior rather than nominal trajectory imitation alone. For quadruped users, \textit{Bypass} sequences represent obstacle-aware detours, while \textit{Blocked} routes are represented through truncated or stop-oriented action sequences. For drones, expert actions reflect a nominal flight-height assumption. Low ground obstacles may be handled by near-straight motion toward the goal, whereas overhead hazards or route-obstructing structures induce detour or stop-like sequences.

\subsection{User-Conditioned Tasks and Evaluation}

Different user embodiments require different navigation interfaces. Unified navigation policies can accept multiple goal modalities~\cite{hirose2026omnivla}, yet for embodied agents such as drones~\cite{xu2026aerialvla, lee2025citynav, li2025skyvln} and quadruped robots~\cite{cheng2024navila, castro2025vamos, tong2024quart}, navigation can be expressed through low-level control signals that determine future pose changes. For humanoid~\cite{pan2025lookout, cheng2024navila}, BVI~\cite{gao2025wearable, joshi2024ai}, and wheelchair users~\cite{wang2024intelligent, xu2025conav}, navigation is more naturally expressed as high-level guidance that communicates safe and feasible route-following behavior. VLAN adopts two user-conditioned task settings: 

\noindent\textbf{Action Prediction Navigation.}
For action-prediction users, the model receives the first three keyframes, the first two actions, the target location, the final keyframe, and the target user embodiment. The model predicts a future sequence of low-level navigation actions
\begin{equation}
\hat{\mathbf{a}} = \{\hat{a}_{t+1}, \hat{a}_{t+2}, \dots, \hat{a}_{t+H}\},
\end{equation}
where each $\hat{a}_{t}$ is represented as $(\Delta x, \Delta z, \Delta \theta)$. These actions can be applied to quadruped robots and drones, whose navigation behavior can be evaluated through offline rollout execution.

\input{Tables/metric}

\noindent\textbf{Language-Guided Navigation.}
For humanoid, BVI, and wheelchair users, the model produces user-specific navigation guidance in natural language rather than executable control actions. These outputs follow distinct conventions and safety priorities. Humanoid guidance uses distance-based and side-based route descriptions. BVI guidance uses steps rather than meters, clock directions rather than left-right language, and explicit risk communication~\cite{10.1145/3597638.3608955}. Wheelchair guidance emphasizes route feasibility, clearance, and surface stability rather than obstacle avoidance alone~\cite{Kasemsuppakorn04052015}. These targets are not stylistic variants of a single template, but distinct user-conditioned guidance forms.

\noindent\textbf{Evaluation protocol} follows the two task settings as in Table~\ref{tab:metric_definitions}. Action Prediction Navigation is evaluated with rollout-based metrics for goal completion, collision avoidance, and path efficiency. Language-Guided Navigation is evaluated with user-conditioned semantic correctness and structured instruction similarity. The required constraints differ by user: humanoid guidance emphasizes accessibility, obstacle semantics, distance, and detour-side correctness. BVI guidance requires step consistency, clock-direction accuracy, risk communication, blocked-route compliance, and non-visual style, while wheelchair guidance emphasizes feasibility, clearance, and blocked-route compliance.

%% file: Tables/0_scene_taxonomy.tex
\begin{wraptable}{r}{0.32\linewidth}
    \vspace{-1.0em}
    \centering
    \small
    \renewcommand{\arraystretch}{1.0}
    \setlength{\tabcolsep}{1.0mm}
    \caption{\textbf{Scene diversity statistics.} Distribution of 1,000 classified scenes across lighting, scene type, path geometry, and environmental condition.}
    \label{tab:scene_taxonomy_summary}
    \begin{tabular}{llc}
    \toprule
    \textbf{Dimension} & \textbf{Category} & \textbf{Count} \\
    \midrule
    \multirow{4}{*}{Lighting}
    & Daytime    & 474 \\
    & Dusk       & 131 \\
    & Nighttime  & 251 \\
    & Low Vision & 144 \\
    \midrule
    \multirow{4}{*}{Scene}
    & Urban      & 366 \\
    & Suburban   & 311 \\
    & Rural      & 273 \\
    & Indoor     & 50   \\
    \midrule
    \multirow{4}{*}{Geometry}
    & Wide        & 399 \\
    & Narrow      & 158 \\
    & Open        & 284 \\
    & Constrained & 159 \\
    \midrule
    \multirow{4}{*}{Env. Cond.}
    & Sunny      & 563 \\
    & Snowy      & 134 \\
    & Rainy      & 42  \\
    & Wet Ground & 261 \\
    \bottomrule
    \end{tabular}
    \vspace{-1.0em}
\end{wraptable}

%% file: Tables/metric.tex
\begin{table*}[t] \label{tab:metric}
\centering
\caption{\textbf{Task-aligned evaluation metrics.}
Action Prediction Navigation is evaluated using execution-oriented metrics over predicted motion rollouts, while Language-Guided Navigation is evaluated using user-conditioned semantic success and structured instruction similarity.
}
\label{tab:metric_definitions}
\small
\setlength{\tabcolsep}{4pt}
\renewcommand{\arraystretch}{1.15}
\begin{tabular}{p{0.10\linewidth}p{0.29\linewidth}p{0.54\linewidth}}
\toprule
\textbf{Metric} & \textbf{Formula} & \textbf{Interpretation} \\
\midrule

$\mathrm{SR}_{\mathrm{act}}$ &
$\displaystyle
\frac{1}{N}\sum_{i=1}^{N}\mathbb{I}\!\left[g^{(i)} = 1 \land c^{(i)} = 1\right]
$ &
Primary success criterion for Action Prediction Navigation. A rollout is successful only if it reaches the goal region and remains collision-free. \\

$\mathrm{GR}$ &
$\displaystyle
\frac{1}{N}\sum_{i=1}^{N} g^{(i)}
$ &
Fraction of predicted rollouts that reach the target region, regardless of collision status. \\

$\mathrm{CF}$ &
$\displaystyle
\frac{1}{N}\sum_{i=1}^{N} c^{(i)}
$ &
Fraction of predicted rollouts that remain collision-free throughout execution. \\

$\mathrm{SPL}$ &
$\displaystyle
\frac{1}{N}\sum_{i=1}^{N}
g^{(i)}
\frac{l_i^{\ast}}{\max(l_i^{\ast},\hat{l}_i)}
$ &
Path-efficiency metric for successful goal-reaching rollouts, comparing predicted path length $\hat{l}_i$ against reference path length $l_i^{\ast}$. \\

$\mathrm{SR}_{\mathrm{lan}}^{u}$ &
$\displaystyle
\frac{1}{N_u}\sum_{i=1}^{N_u}
\mathbb{I}\!\left[
\bigwedge_{k\in\mathcal{K}_u} m_k^{(i,u)} = 1
\right]
$ &
Strict semantic success rate for user type $u \in \{\text{humanoid}, \text{BVI}, \text{wheelchair}\}$. A prediction is successful only if all required user-specific semantic constraints are satisfied. \\

$\mathrm{IAS}^{u}$ &
$\displaystyle
\frac{1}{N_u}\sum_{i=1}^{N_u}
\sum_{k\in\mathcal{K}_u} w_k^{(u)} s_k^{(i,u)}
$ &
Structured similarity between the predicted instruction and the canonical target for user type $u$. 
\\

\bottomrule
\end{tabular}
\vspace{4pt}

\raggedright\scriptsize{
Here, $g^{(i)} \in \{0,1\}$ indicates goal reaching, $c^{(i)} \in \{0,1\}$ indicates collision-free execution, $l_i^{\ast}$ is the reference path length, and $\hat{l}_i$ is the predicted rollout length. For Language-Guided Navigation, $
\mathcal{K}_u$ denotes the set of required semantic constraints for user type $u$, $m_k^{(i,u)} \in \{0,1\}$ is the binary correctness term for constraint $k$, $s_k^{(i,u)} \in [0,1]$ is its continuous similarity term, and $\sum_{k \in
\mathcal{K}_u} w_k^{(u)} = 1$.
}
\end{table*}

%% file: Sections/4_experiments.tex
\section{Experiments}

\subsection{Experimental Setup}

The navigation evaluations are conducted on the VLAN test split, which contains 100 scenes and 1,931 trajectories per user setting. The benchmark covers five user types: \textit{humanoid}, \textit{BVI}, \textit{wheelchair user}, \textit{drone}, and \textit{quadruped robot}. Following Sec.~\ref{sec:dataset}, these users are grouped into two task families: Action Prediction Navigation for drone and quadruped users (Sec.~\ref{sec4.2}), Language-Guided Navigation for humanoid, BVI, and wheelchair users (Sec.~\ref{sec4.3}).

\textbf{Baselines.} Seven MLLMs have completed evaluation on all five user settings: \textbf{Qwen2.5-VL-3B-Instruct}~\cite{bai1others}, \textbf{Qwen2.5-VL-7B-Instruct}, \textbf{Qwen3-VL-8B-Instruct}~\cite{bai2025qwen3}, \textbf{InternVL3.5-8B}~\cite{wang2025internvl3}, \textbf{Idefics3-8B-Llama3}~\cite{laurenccon2024building}, \textbf{Molmo2-8B}~\cite{clark2026molmo2}, and \textbf{Llama-3.2-11B-Vision-Instruct}. 

For Language-Guided Navigation, we report $\mathrm{SR}_{\mathrm{lan}}$ and $\mathrm{IAS}$. For Action Prediction Navigation, we report $\mathrm{SR}_{\mathrm{act}}$, $\mathrm{GR}$, $\mathrm{CF}$, and $\mathrm{SPL}$. The results are shown in Tables~\ref{tab:simplified_action_results} and~\ref{tab:simplified_language_results}.

\subsection{Action Prediction Navigation Results}\label{sec4.2}

\input{Tables/eval_actions}

\textbf{Drone.}
As presented in Table~\ref{tab:simplified_action_results}, InternVL3.5-8B achieves the strongest overall drone performance, with the best $\mathrm{SR}_{\mathrm{act}}$, $\mathrm{GR}$, and $\mathrm{SPL}$. Qwen3-VL-8B-Instruct is a close second and remains highly competitive across all drone
metrics. Qwen2.5-VL-7B-Instruct also performs well, indicating that medium-scale models can already support strong action-conditioned aerial navigation when the output space is explicit and geometric. In contrast, Qwen2.5-VL-3B-Instruct
achieves the highest collision-free rate but extremely low success and path efficiency. This pattern indicates conservative behavior: the model often avoids collisions, but fails to reach the goal. Llama-3.2-11B-Vision-Instruct and
Molmo2-8B show similar underperformance, suggesting that strong language-oriented capability does not automatically transfer to precise action rollout quality, consistent with findings that MLLMs remain bottlenecked by visual grounding and action refinement even when given a valid plan~\cite{shi2026spaceact}.

\textbf{Quadruped Robot.}
Qwen3-VL-8B-Instruct is the best model for quadruped navigation, achieving the highest action success rate, goal-reaching rate, and SPL. InternVL3.5-8B remains competitive and is again the most balanced cross-user model, but Qwen3-VL-
8B-Instruct is stronger in this specific action regime. Qwen2.5-VL-7B-Instruct also performs well, while Idefics3-8B-Llama3 is moderate. As with the drone setting, Qwen2.5-VL-3B-Instruct achieves extremely high collision-free execution
but very low success, showing a strong safe-but-inactive tendency. Llama-3.2-11B-Vision-Instruct is particularly weak on the quadruped task, which further confirms that low-level embodied action prediction is a distinct challenge from language-guided route generation.

\input{Tables/eval_language}

\subsection{Language-Guided Navigation Results}\label{sec4.3}

Table~\ref{tab:simplified_language_results} reports humanoid, BVI, and wheelchair results. The user types expose different failure modes.

\textbf{Humanoid.}
Humanoid navigation is closest to generic language-guided route following, yet it still requires correct accessibility prediction, detour-side reasoning, and quantitative guidance. Llama-3.2-11B-Vision-Instruct achieves the best performance, obtaining the highest $\mathrm{SR}_{\mathrm{lan}}$ and $\mathrm{IAS}$, with Molmo2-8B as a close second. Qwen2.5-VL-7B-Instruct achieves a relatively strong $\mathrm{IAS}$ but a much lower success rate, suggesting that partial semantic overlap does not ensure fully correct user-conditioned guidance. This gap indicates that fluent route descriptions are insufficient when accessibility semantics or detour direction is not preserved.

\textbf{BVI.}
Non-visual navigation guidance is substantially more challenging than generic route description, as it requires step-based distances, clock-direction cues, and explicit risk communication. Qwen3-VL-8B-Instruct achieves the highest success rate, while Molmo2-8B obtains the highest $\mathrm{IAS}$, suggesting that Molmo2-8B better matches the target instruction style, whereas Qwen3-VL-8B-Instruct more reliably satisfies safety-critical semantic constraints. InternVL3.5-8B remains competitive on both metrics, indicating robust user-specific semantic alignment. In contrast, Idefics3-8B-Llama3 performs weakly, showing that generic instruction quality does not necessarily transfer to safety-aware non-visual navigation.

\textbf{Wheelchair.}
For wheelchair users, the central requirement is route feasibility under width, clearance, and surface-stability constraints. InternVL3.5-8B achieves the highest success rate, while Molmo2-8B obtains the highest $\mathrm{IAS}$. The gap between strict success and alignment again suggests that producing plausible guidance is easier than satisfying accessibility constraints precisely. Qwen2.5-VL-3B-Instruct also performs strongly, whereas Qwen3-VL-8B-Instruct drops substantially compared with its BVI and action-prediction results. This indicates that wheelchair-oriented guidance depends on precise feasibility reasoning rather than general route-description quality alone.

\subsection{Overall Comparison Across Users}

Per-user success rates are not directly comparable across the two task families. To address this, we introduce \textit{Accessible Success Rate} (\textbf{AccSR}), a balanced aggregate score for overall VLAN performance. We first compute the average success rate within each task family:
\begin{equation}
\mathrm{SR}_{\mathrm{lan}}^{\mathrm{avg}}
=
\frac{
\mathrm{SR}_{\mathrm{lan}}^{\mathrm{humanoid}}
+
\mathrm{SR}_{\mathrm{lan}}^{\mathrm{BVI}}
+
\mathrm{SR}_{\mathrm{lan}}^{\mathrm{wheelchair}}
}{3},
\mathrm{SR}_{\mathrm{act}}^{\mathrm{avg}}
=
\frac{
\mathrm{SR}_{\mathrm{act}}^{\mathrm{drone}}
+
\mathrm{SR}_{\mathrm{act}}^{\mathrm{quadruped}}
}{2}.
\end{equation}
We then define AccSR as the harmonic mean of the two family-level scores:
\begin{equation}
\mathrm{AccSR}
=
\frac{
2 \, \mathrm{SR}_{\mathrm{lan}}^{\mathrm{avg}} \, \mathrm{SR}_{\mathrm{act}}^{\mathrm{avg}}
}{
\mathrm{SR}_{\mathrm{lan}}^{\mathrm{avg}} + \mathrm{SR}_{\mathrm{act}}^{\mathrm{avg}}
}.
\end{equation}

\input{Tables/eval_accessible}

Table~\ref{tab:accessible_crossuser} reports the cross-user comparison. InternVL3.5-8B achieves the highest AccSR, indicating the strongest overall balance across language-guided and action-prediction settings. Qwen3-VL-8B-Instruct performs well on action prediction, but its lower wheelchair-user performance reduces its balanced score. Qwen2.5-VL-7B-Instruct ranks third with a relatively even profile, although it remains below the top models in both task families.

Idefics3-8B-Llama3 performs better for action-based users, whereas Molmo2-8B is stronger for language-guided users. Llama-3.2-11B-Vision-Instruct and Qwen2.5-VL-3B-Instruct receive lower AccSR despite strong performance on selected users, suggesting limited transfer across heterogeneous accessibility requirements. These results show that AccSR captures broad user-conditioned robustness more effectively than isolated per-user scores or an unbalanced overall average.

\subsection{Perception versus Planning Failure Analysis}
\label{subsec:failure}

Aggregate success rates do not indicate which stage of the pipeline a model fails at. To separate perception from planning, we jointly consider two per-trajectory signals: whether the model's language output names the correct obstacle category, which we term a perception success $P^{+}$, and whether its predicted rollout succeeds under $\mathrm{SR}_{\mathrm{act}}$, which we term an action success $A^{+}$. Table~\ref{tab:perception_action} reports the resulting four-way partition over all evaluated trajectories.

\input{Tables/failure_perception_action}

The partition distinguishes cases that aggregate scores cannot separate. For Qwen2.5-VL-3B-Instruct and Molmo2-8B, the two models with the lowest $\mathrm{SR}_{\mathrm{act}}$, failures are dominated by the $P^{-}A^{-}$ cell (63.4\% and 61.4\% respectively), indicating that incorrect obstacle grounding coincides with unsuccessful action in most cases. Qwen2.5-VL-3B-Instruct nevertheless exhibits the largest $P^{+}A^{-}$ proportion among all models (23.3\%), so for nearly a quarter of its trajectories the obstacle is identified correctly and the predicted action still fails. This cell provides the quantitative counterpart to the safe-but-inactive behaviour discussed in Sec.~\ref{sec4.2}, and consistent with evidence that fine-grained visual perception remains a distinct bottleneck for MLLMs~\cite{guan2026perceptrl}. At the other end, Qwen3-VL-8B-Instruct and InternVL3.5-8B concentrate the largest share of trajectories in $P^{+}A^{+}$ (28.9\% and 26.5\%), consistent with their higher aggregate success. Idefics3-8B-Llama3 presents a contrasting profile: its $P^{-}A^{+}$ proportion is the highest of all models (40.4\%), meaning it reaches the goal while misnaming the obstacle, so this subset of its successes is not attributable to accessibility reasoning.

Two qualifications follow from the construction of these cells. A correct identification followed by an unsuccessful action does not by itself isolate action generation, since spatial grounding and user conditioning may both contribute within $P^{+}A^{-}$; conversely, $P^{-}A^{-}$ bounds the perception contribution without attributing the remainder. We therefore interpret the four cells as operational outcome categories rather than as a fully disentangled causal decomposition.

\subsection{Sim-to-Real Evaluation}
To assess whether the benchmark transfers beyond simulation, we additionally conduct a sim-to-real evaluation in a real deployment setting. We further collect 30 real-world trajectories from physical scenes and evaluate the same five user-conditioned tasks. As presented in Table~\ref{tab:accessible_sr}, the performance of methods are lower yet comparable with the ones in VLAN, and the relative ordering of models is largely preserved. Because this check covers only 30 trajectories, we treat it as preliminary evidence that VLAN performance correlates with real-world performance rather than as a definitive real-world evaluation, and we return to this point in the Limitations of Appendix~\ref{app:limitations}.

\input{Tables/eval_sim2real}

%% file: Tables/eval_actions.tex
\begin{table*}[ht]
\centering
\caption{\textbf{Action-prediction navigation results in VLAN.}
We report $\mathrm{SR}_{\mathrm{act}}$, $\mathrm{GR}$, $\mathrm{CF}$, and $\mathrm{SPL}$ for the two action-prediction users (Drone and Quadruped Robot). Only models with completed evaluation on all five user settings are included.
}
\label{tab:simplified_action_results}
\small
\setlength{\tabcolsep}{12pt}
\renewcommand{\arraystretch}{1.0}
\begin{adjustbox}{width=0.99\linewidth}
\begin{tabular}{l c c c c c}
\toprule
\textbf{Model} & \textbf{Params} & $\mathbf{SR_{\mathrm{act}}}$ $\uparrow$ & $\mathbf{GR}$ $\uparrow$ & $\mathbf{CF}$ $\uparrow$ & $\mathbf{SPL}$ $\uparrow$ \\
\midrule

\rowcolor{gray!20}
\multicolumn{6}{l}{\textit{(a) Drone}} \\
\midrule 
Qwen2.5-VL-3B-Instruct & 3B & 0.133 & 0.131 & \textbf{0.947} & 0.133 \\
Qwen2.5-VL-7B-Instruct & 7B & 0.571 & 0.608 & 0.732 & 0.569 \\
Qwen3-VL-8B-Instruct & 8B & \underline{0.622} & \underline{0.645} & 0.758 & \underline{0.597} \\
InternVL3.5-8B & 8B & \textbf{0.642} & \textbf{0.675} & 0.794 & \textbf{0.621} \\
Idefics3-8B-Llama3 & 8B & 0.554 & 0.579 & 0.648 & 0.538 \\
Molmo2-8B & 8B & 0.289 & 0.312 & 0.793 & 0.284 \\
Llama-3.2-11B-Vision-Instruct & 11B & 0.247 & 0.260 & \underline{0.799} & 0.246 \\
\midrule

\rowcolor{gray!20}
\multicolumn{6}{l}{\textit{(b) Quadruped Robot}} \\
\midrule 
Qwen2.5-VL-3B-Instruct & 3B & 0.053 & 0.054 & \textbf{0.983} & 0.053 \\
Qwen2.5-VL-7B-Instruct & 7B & 0.384 & \underline{0.542} & 0.741 & \underline{0.383} \\
Qwen3-VL-8B-Instruct & 8B & \textbf{0.474} & \textbf{0.584} & 0.770 & \textbf{0.458} \\
InternVL3.5-8B & 8B & \underline{0.393} & 0.530 & 0.808 & 0.375 \\
Idefics3-8B-Llama3 & 8B & 0.308 & 0.467 & 0.650 & 0.299 \\
Molmo2-8B & 8B & 0.147 & 0.253 & 0.830 & 0.143 \\
Llama-3.2-11B-Vision-Instruct & 11B & 0.048 & 0.108 & \underline{0.893} & 0.048 \\
\bottomrule
\end{tabular}
\end{adjustbox}
\end{table*}

%% file: Tables/eval_language.tex
\begin{wraptable}{r}{0.58\linewidth}
\vspace{-0.8em}
\centering
\caption{\textbf{Language-guided navigation results.}
We report $\mathrm{SR}_{\mathrm{lan}}$ and $\mathrm{IAS}$ for the three language-guided users. Only models with completed evaluation on all five user settings are included.
}
\label{tab:simplified_language_results}
\small
\setlength{\tabcolsep}{2.8pt}
\renewcommand{\arraystretch}{0.92}
\begin{tabular}{p{0.51\linewidth} c c c}
\toprule
\textbf{Model} & \textbf{Params} & $\mathbf{SR_{\mathrm{lan}}}$ $\uparrow$ & $\mathbf{IAS}$ $\uparrow$ \\
\midrule

\rowcolor{gray!20}
\multicolumn{4}{l}{\textit{(a) Humanoid}} \\
\midrule
Qwen2.5-VL-3B-Instruct & 3B & 0.483 & 0.752 \\
Qwen2.5-VL-7B-Instruct & 7B & 0.275 & 0.781 \\
Qwen3-VL-8B-Instruct & 8B & 0.389 & 0.742 \\
InternVL3.5-8B & 8B & 0.365 & 0.738 \\
Idefics3-8B-Llama3 & 8B & 0.438 & 0.750 \\
Molmo2-8B & 8B & \underline{0.523} & \underline{0.840} \\
Llama-3.2-11B-Vision-Instruct & 11B & \textbf{0.558} & \textbf{0.846} \\
\midrule

\rowcolor{gray!20}
\multicolumn{4}{l}{\textit{(b) BVI}} \\
\midrule
Qwen2.5-VL-3B-Instruct & 3B & 0.265 & 0.728 \\
Qwen2.5-VL-7B-Instruct & 7B & 0.276 & 0.719 \\
Qwen3-VL-8B-Instruct & 8B & \textbf{0.436} & \underline{0.767} \\
InternVL3.5-8B & 8B & \underline{0.375 }& 0.774 \\
Idefics3-8B-Llama3 & 8B & 0.118 & 0.576 \\
Molmo2-8B & 8B & 0.131 & \textbf{0.781} \\
Llama-3.2-11B-Vision-Instruct & 11B & 0.183 & 0.723 \\
\midrule

\rowcolor{gray!20}
\multicolumn{4}{l}{\textit{(c) Wheelchair}} \\
\midrule
Qwen2.5-VL-3B-Instruct & 3B & \underline{0.332} & 0.645 \\
Qwen2.5-VL-7B-Instruct & 7B & 0.208 & 0.672 \\
Qwen3-VL-8B-Instruct & 8B & 0.092 & 0.629 \\
InternVL3.5-8B & 8B & \textbf{0.347} & \underline{0.684} \\
Idefics3-8B-Llama3 & 8B & 0.004 & 0.576 \\
Molmo2-8B & 8B & 0.289 & \textbf{0.748} \\
Llama-3.2-11B-Vision-Instruct & 11B & 0.075 & 0.644 \\
\bottomrule
\end{tabular}
\vspace{-0.8em}
\end{wraptable}

%% file: Tables/eval_accessible.tex
  \begin{table}[t]
  \centering
  \caption{\textbf{Comparison cross users.} 
  We first average success rates within the language-guided and action-prediction task families, and then compute their harmonic mean as a balanced overall score. 
  }
  \label{tab:accessible_crossuser}
  \small
  \setlength{\tabcolsep}{6pt}
  \renewcommand{\arraystretch}{0.98}
  
  \begin{adjustbox}{width=\linewidth}
  \begin{tabular}{p{0.42\linewidth} c c c c}
  \toprule
  \textbf{Model} & \textbf{Params} & $\mathbf{SR_{\mathrm{lan}}^{avg}}$ $\uparrow$ & $\mathbf{SR_{\mathrm{act}}^{avg}}$ $\uparrow$ & $\mathbf{AccSR}$ $\uparrow$ \\
  \midrule
  Qwen2.5-VL-3B-Instruct & 3B & \underline{0.360} & 0.093 & 0.147 \\
  Qwen2.5-VL-7B-Instruct & 7B & 0.253 & 0.477 & 0.331 \\
  Qwen3-VL-8B-Instruct & 8B & 0.306 & \textbf{0.548} & \underline{0.392} \\
  InternVL3.5-8B & 8B & \textbf{0.363} & \underline{0.517} & \textbf{0.426} \\
  Idefics3-8B-Llama3 & 8B & 0.186 & 0.431 & 0.260 \\
  Molmo2-8B & 8B & 0.315 & 0.218 & 0.258 \\
  Llama-3.2-11B-Vision-Instruct & 11B & 0.272 & 0.148 & 0.191 \\
  
  \bottomrule
  \end{tabular}
  \end{adjustbox}
  \end{table}

%% file: Tables/failure_perception_action.tex
\begin{table*}[ht]
\centering
\caption{\textbf{Cross-analysis of perception correctness and action generation success} (\%).
$P^{+}$/$P^{-}$ denote correct/incorrect obstacle perception in the language output, and $A^{+}$/$A^{-}$ denote successful/failed action prediction. $P^{+}A^{-}$ represents pure planning failure.}
\label{tab:perception_action}
\small
\setlength{\tabcolsep}{10pt}
\renewcommand{\arraystretch}{1.0}
 \begin{adjustbox}{width=\linewidth}
\begin{tabular}{l c c c c}
\toprule
\textbf{Model} & $\mathbf{P^{+}A^{+}}$ & $\mathbf{P^{+}A^{-}}$ & $\mathbf{P^{-}A^{+}}$ & $\mathbf{P^{-}A^{-}}$ \\
\midrule
Qwen2.5-VL-3B-Instruct        & 4.8  & 23.3 & 8.5  & 63.4 \\
Qwen2.5-VL-7B-Instruct        & 17.3 & 12.8 & 39.7 & 30.1 \\
Qwen3-VL-8B-Instruct          & 28.9 & 16.4 & 33.2 & 21.4 \\
InternVL3.5-8B                & 26.5 & 12.3 & 37.6 & 23.6 \\
Idefics3-8B-Llama3            & 14.9 & 10.4 & 40.4 & 34.2 \\
Molmo2-8B                     & 4.8  & 9.7  & 24.1 & 61.4 \\
Llama-3.2-11B-Vision-Instruct & 4.5 & 16.7 & 20.1 & 58.7 \\
\bottomrule
\end{tabular}
\end{adjustbox}
\end{table*}

%% file: Tables/eval_sim2real.tex
  \begin{table}[ht]
  \centering
  \caption{\textbf{Sim-to-real evaluation on 30 real-world navigation trajectories. }
  We report task-specific success rates and the overall Accessible Success Rate (AccSR) across the five user-conditioned tasks.
  }
  \label{tab:accessible_sr}
  \small
  \setlength{\tabcolsep}{4pt}
  \renewcommand{\arraystretch}{0.98}
  \begin{adjustbox}{width=\linewidth}
  \begin{tabular}{p{0.42\linewidth} c c c c}
  \toprule
  \textbf{Model} & \textbf{Params} & $\mathbf{SR_{\mathrm{lan}}^{avg}}$ $\uparrow$ & $\mathbf{SR_{\mathrm{act}}^{avg}}$ $\uparrow$ & $\mathbf{AccSR}$ $\uparrow$ \\
  \midrule
  Qwen2.5-VL-3B-Instruct & 3B & \textbf{0.233} & 0.083 & 0.123 \\
  Qwen2.5-VL-7B-Instruct & 7B & 0.189 & 0.467 & 0.269 \\
  Qwen3-VL-8B-Instruct & 8B & \underline{0.200} & \underline{0.483} & \underline{0.283} \\
  InternVL3.5-8B & 8B & \underline{0.200} & \textbf{0.617} & \textbf{0.302} \\
  Idefics3-8B-Llama3 & 8B & 0.111 & 0.417 & 0.175 \\
  Molmo2-8B & 8B & 0.133 & 0.350 & 0.193 \\
  Llama-3.2-11B-Vision-Instruct & 11B & 0.133 & 0.217 & 0.165 \\
  
  \bottomrule
  \end{tabular}
  \end{adjustbox}
  \end{table}

%% file: Sections/5_conclusion.tex
\section{Conclusion}\label{sec:conclusion}
We propose \textbf{Vision-Language Accessible Navigation (VLAN)}, a benchmark for the first time to evaluate navigation under user-specific \textit{accessibility} constraints. Unlike existing navigation benchmarks that assume a fixed embodiment, VLAN captures how mobility changes across humanoid and quadruped robots, wheeled agents, and people with visual or mobility impairments. With diverse scenes, trajectories, keyframes, obstacle annotations, and accessibility labels, VLAN enables evaluation beyond goal reaching, including collision awareness and accessibility alignment. Extensive experiments show that current vision-language models still struggle to adapt path selection, obstacle relevance, and safety reasoning across different user profiles. These insights highlight the need for accessibility-aware datasets, tasks, and metrics for future embodied AI systems.

%% file: appendix.tex
\clearpage
\appendix
\raggedbottom

\newcommand{\curprot}{Cur}
\newcommand{\burstprot}{Burst}
\newcommand{\histprot}{Hist}
\definecolor{occOthers}{RGB}{0,0,0}
\definecolor{occBarrier}{RGB}{255,120,50}
\definecolor{occBicycle}{RGB}{255,192,203}
\definecolor{occBus}{RGB}{255,255,0}
\definecolor{occCar}{RGB}{0,150,245}
\definecolor{occConstruction}{RGB}{0,255,255}
\definecolor{occMotorcycle}{RGB}{255,127,0}
\definecolor{occPedestrian}{RGB}{255,0,0}
\definecolor{occTrafficCone}{RGB}{255,240,150}
\definecolor{occTrailer}{RGB}{135,60,0}
\definecolor{occTruck}{RGB}{160,32,240}
\definecolor{occDriveable}{RGB}{255,0,255}
\definecolor{occOtherFlat}{RGB}{139,137,137}
\definecolor{occSidewalk}{RGB}{75,0,75}
\definecolor{occTerrain}{RGB}{150,240,80}
\definecolor{occManmade}{RGB}{230,230,250}
\definecolor{occVegetation}{RGB}{0,175,0}
\newcommand{\occbox}[1]{\raisebox{0.15ex}{\textcolor{#1}{\rule{0.85ex}{0.85ex}}}}
\newcommand{\occhead}[2]{\rotatebox[origin=c]{90}{\scriptsize\occbox{#1}\hspace{0.25em}#2\strut}}

\section{Limitations and Broader Impacts}
\label{app:limitations}

In this work, we focus on dataset construction and benchmark evaluation for next-generation accessible navigation. The considered embodiments do not fully cover all user groups, such as people with hearing or cognitive impairments. The current annotations capture common accessibility constraints, but individual users may have more diverse needs, preferences, abilities, and assistive strategies. Future work can extend VLAN by involving broader user groups, incorporating richer multimodal inputs, and developing more personalized user models.

Several design decisions bound what the reported numbers can support, and we state them explicitly here. As detailed in Appendix~\ref{app:drone2d}, the drone task is defined on a fixed-altitude plane, so vertical motion lies outside the definition of the task rather than being a capability that the benchmark omits, which restricts the drone results to fixed-altitude 2D accessibility-conditioned navigation. Accessibility itself is compressed into four decision-oriented categories, which support consistent annotation and measurable evaluation but cannot represent graded risk, personal preference, or capability profiles beyond the five considered embodiments. The sim-to-real study covers only 30 real-world trajectories, so we treat it as preliminary evidence that simulated and real performance agree in ordering rather than as a definitive real-world evaluation. All reported scores come from a single deterministic greedy decoding run, which makes them reproducible but provides no across-run variance, so we report point estimates without error bars; the sole source of sampling variance in this work is the 200-trajectory subset used for the closed-source reference models in Appendix~\ref{app:closed_source}.

These limitations do not diminish the intended use of VLAN. The benchmark aims to support the development of safer and more inclusive navigation systems for diverse users and embodied agents. By emphasizing accessibility-aware evaluation, it may help future AI systems better serve people with visual or mobility impairments. At the same time, such navigation assistance systems should be carefully tested before broader real-world use, since incorrect navigation guidance may create risks for vulnerable users.

\section{Fixed-Altitude 2D Drone Formulation}
\label{app:drone2d}

\subsection{Design Rationale}

VLAN evaluates whether a model adapts its navigation reasoning to embodiment-specific accessibility constraints under identical visual input, rather than whether an agent can execute general 3D flight. Keeping the drone on a fixed-altitude plane isolates accessibility reasoning from flight-control precision: adding vertical degrees of freedom would increase the dimensionality of the output space and compound failure sources across perception, spatial reasoning, user conditioning, and action precision, which would make performance degradation difficult to attribute to a specific cause.

The fixed-altitude abstraction also has precedent in assistive drone guidance. DroneNavigator~\cite{avila2017dronenavigator} uses a near-ground constant-altitude quadcopter with 2D planar guidance to lead visually impaired travellers; Flying Guide Dog~\cite{tan2021flying} performs low-altitude horizontal-plane guidance from walkable-area segmentation; GuideCopter~\cite{huppert2021guidecopter} provides planar haptic guidance through tether-mediated interaction; the Aerial Guide Dog~\cite{zhang2024aerial} evaluates indoor 2D navigation guidance by directional accuracy and path deviation; and terrain-following constant-AGL abstractions~\cite{wubben2022improving} maintain a stable altitude for autonomous UAV missions. VLAN adopts the same constant-altitude operating assumption for the same reason: it targets accessibility-conditioned route selection rather than altitude control.

A further practical consideration motivates sharing the action space across the two action-prediction embodiments. Both the drone and the quadruped robot are represented by $(\Delta x, \Delta z, \Delta\theta)$, so their scores are computed by the same rollout procedure against the same goal tolerance and collision clearance, and remain directly comparable on the same scene. Introducing a vertical dimension for the drone alone would require a separate evaluation protocol and make cross-embodiment comparison less interpretable.

\subsection{Annotation Policy and Scope}

The annotation follows an explicit default setting in which each embodiment is assigned a fixed body height and width, and an obstacle is labelled relevant to a user if and only if it intersects that user's body volume. The quadruped robot has the lowest body profile and is therefore not blocked by shop signs, roadside protruding branches, or protruding traffic signs that lie above its body band. The wheelchair user has a somewhat higher profile, but its constraints arise from wheelchair width and ground traversability rather than from height. BVI and humanoid users are assigned normal pedestrian height. The drone is the embodiment most directly related to obstacle height: it maintains a constant flight altitude above pedestrian level, so all overhead-type obstacles are treated as obstructing its route at that altitude, and its avoidance action space is restricted to the 2D plane.

Under this policy the drone labels, reference actions, and metrics are defined on the fixed-altitude plane by construction. The task is a fixed-altitude 2D navigation task, so vertical motion lies outside its definition rather than being a capability the benchmark omits. Of the 1,931 test trajectories, 1,371 (71.0\%) are labelled \textit{Passable} for the drone because the obstacle is irrelevant at its flight altitude, while the remaining 560 (29.0\%) require either a horizontal detour (\textit{Bypass}, 538 trajectories) or a stop (\textit{Blocked}, 22 trajectories). The 560 trajectories are the subset for which an altitude-capable policy could in principle yield a different label or reference action; a definitive answer would require re-annotating obstacle clearance heights and reference altitudes under a 3D policy, which we leave to future work.

\subsection{Differential Analysis}

A direct measurement of how labels would change under a 3D policy is not available, but the composition of the test split can be probed indirectly. If the 560 constrained trajectories encode drone-specific difficulty rather than general scene difficulty, then removing them should benefit the drone substantially more than the quadruped robot, since overhead obstacles do not obstruct a ground agent under the same annotation policy.

Table~\ref{tab:drone_differential} reports this comparison. Across all seven models the difference-in-differences $\Delta\Delta$ is positive, ranging from $+0.035$ to $+0.098$: drone success improves by $+0.034$ to $+0.100$ on the unconstrained subset, while quadruped success changes by only $-0.012$ to $+0.010$. Model rankings and failure patterns are preserved on the unconstrained subset. The constrained trajectories therefore introduce difficulty that is specific to an aerial embodiment, which is the behaviour expected if the 2D formulation captures embodiment-conditioned accessibility rather than a generic scene-level confound.

\input{Tables/drone_differential}

We note what this analysis does and does not establish. It shows that the constrained subset is disproportionately hard \emph{for the drone}, which supports the claim that VLAN encodes embodiment-specific accessibility constraints. It does not establish that the fixed-altitude labels coincide with the labels a 3D policy would assign, because no altitude-capable annotation exists to compare against. The 29.0\% figure should therefore be read as a statement about the scope of the 2D policy, not as an estimate of annotation error.





\section{Dataset Visualization}
\label{app:visualization}

This section provides qualitative views of the scene and trajectory diversity that underlie the statistics in Table~\ref{tab:scene_taxonomy_summary}.

\subsection{Scene-Level Diversity}

\begin{figure}[H]
\centering
\includegraphics[width=1.0\textwidth]{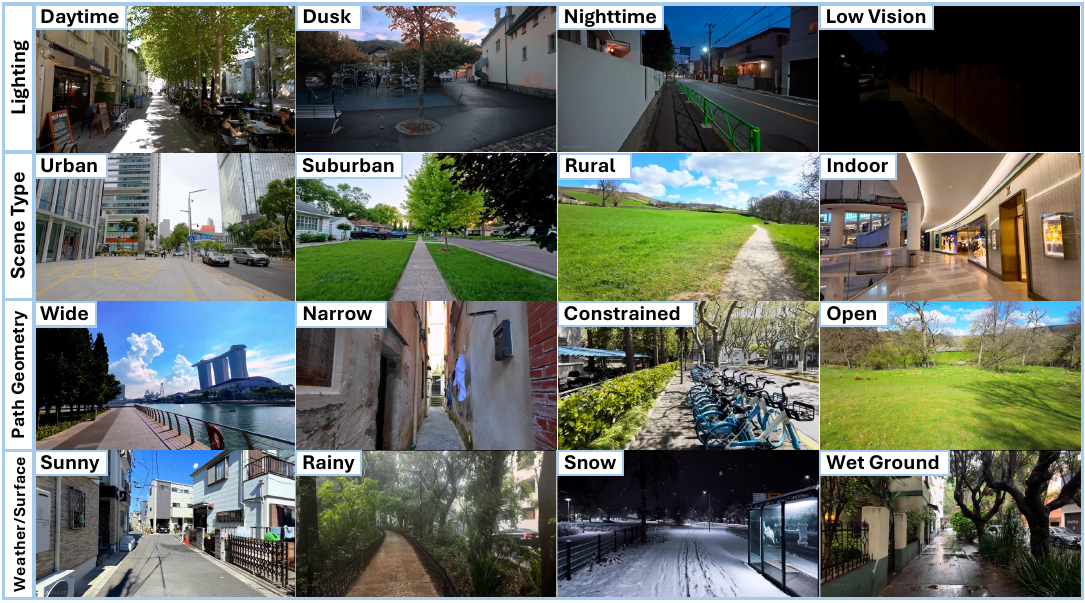}
\caption{\textbf{Scene diversity visualization.} Representative scenes along the four taxonomy axes of Table~\ref{tab:scene_taxonomy_summary}: lighting (daytime, dusk, nighttime, low vision), scene type (urban, suburban, rural, indoor), path geometry (wide, narrow, constrained, open), and weather or surface condition (sunny, rainy, snow, wet ground).}
\label{fig:apen1}
\end{figure}

Fig.~\ref{fig:apen1} organizes representative scenes along the four taxonomy axes used in Table~\ref{tab:scene_taxonomy_summary}. Each row corresponds to one dimension---lighting, scene type, path geometry, and weather or surface condition---and each column shows a scene drawn from the corresponding category. The lighting row contrasts captures of comparable geometry under daytime, dusk, nighttime, and low-vision conditions, which is the axis along which Table~\ref{tab:results_by_lighting} reports the largest variation in AccSR. The scene-type and path-geometry rows illustrate the range from wide open routes to narrow constrained passages, and the weather row isolates surface condition at otherwise similar viewpoints. Because VLAN reconstructs rather than re-collects each scene, the same underlying geometry can appear under several of these conditions, which is why the taxonomy axes are not independent of one another in the released data.

\subsection{Edited Trajectories and Derived Expert Actions}

\begin{figure}[H]
\centering
\includegraphics[width=1.0\textwidth]{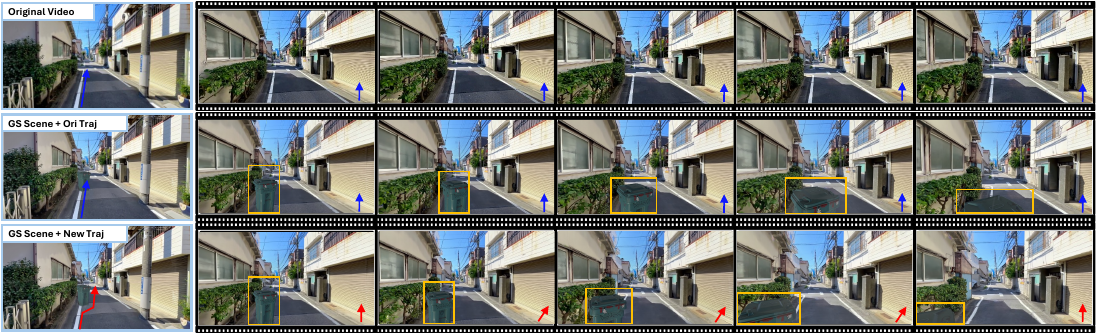}
\caption{\textbf{Edited trajectory and derived expert actions for scene \#0052.} Rows from top to bottom: the original egocentric video frame; the Gaussian-splat reconstruction with the original captured trajectory; and the same reconstruction with a newly generated trajectory together with the expert actions derived from it.}
\label{fig:apen3}
\end{figure}

\begin{figure}[H]
\centering
\includegraphics[width=1.0\textwidth]{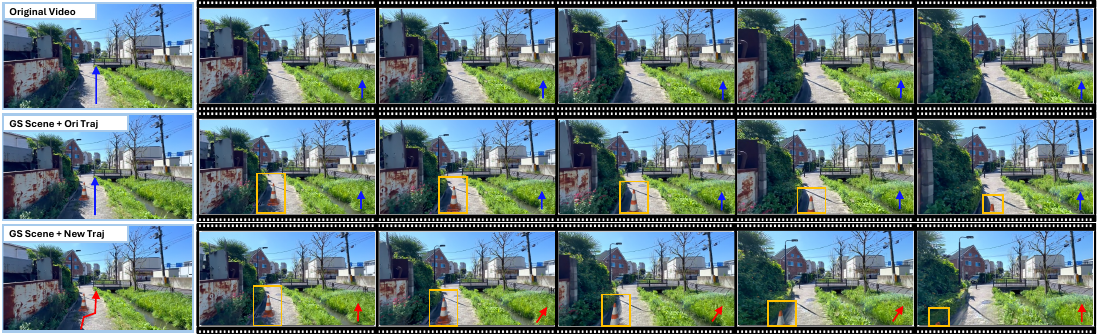}
\caption{\textbf{Edited trajectory and derived expert actions for scene \#0061.} Rows from top to bottom: the original egocentric video frame; the Gaussian-splat reconstruction with the original captured trajectory; and the same reconstruction with a newly generated trajectory together with the expert actions derived from it.}
\label{fig:apen2}
\end{figure}

Figs.~\ref{fig:apen3} and~\ref{fig:apen2} illustrate the editing stage of the pipeline for two scenes. Each figure is organized in three rows. The first row shows the original egocentric video frame, which provides the photorealistic appearance inherited from the reconstruction. The second row shows the Gaussian-splat reconstruction with the original captured trajectory overlaid. The third row shows the same reconstruction after editing, with a newly generated trajectory and the expert actions derived from it.

Comparing the second and third rows makes the role of the editing stage concrete: the visual scene, obstacle layout, and camera viewpoints are preserved, while the route itself is replaced by one generated under the annotation policy of Sec.~\ref{subsec:user_conditioned_trajectories}. This is the mechanism that lets a single reconstructed scene yield multiple navigation instances with different accessibility implications, which is the scaling argument made in Sec.~\ref{subsec:dataset_construction}. Because the underlying geometry is shared across the generated variants, differences in the derived labels and actions are attributable to the route and obstacle configuration rather than to scene appearance.

\section{Additional Experiment Results}
\label{app:results}

This section reports analyses that support the main results but are not essential to them: performance broken down by scene attributes, per-step action statistics, a rollout-level decomposition of failures, cross-user prediction consistency, and evaluation of closed-source reference models.

\subsection{Results by Lighting Condition}

\input{Tables/results_by_lighting}

Table~\ref{tab:results_by_lighting} shows that the relative ranking of models is broadly stable across lighting conditions, but performance degrades under more challenging settings such as low-vision scenes and dusk. InternVL3.5-8B is the most consistent model across all four lighting categories, achieving the best performance in daytime, low-vision, and dusk scenes, while Qwen3-VL-8B-Instruct is slightly stronger in nighttime scenes. In contrast, weaker models such as Qwen2.5-VL-3B-Instruct and Llama-3.2-11B-Vision-Instruct show a larger performance drop under degraded visibility. These results indicate that lighting variation remains an important source of difficulty even when user-conditioned accessibility reasoning is modeled explicitly.

\subsection{Results by Path Geometry}

\input{Tables/results_by_geo}

Table~\ref{tab:results_by_path_geometry} shows that path geometry also has a clear impact on user-conditioned navigation performance. InternVL3.5-8B remains the most stable model across all four geometry categories and achieves the best AccSR in wide, open, narrow, and constrained paths. In general, constrained and narrow scenes are more challenging than wide or open scenes, especially for models with weaker cross-user robustness. These results indicate that accessible navigation performance depends not only on user type and obstacle semantics, but also on the spatial structure of the route itself, and that it depends strongly on the interaction between user type, route semantics, and obstacle-conditioned decision making.

\subsection{Per-Step Action Statistics}

Aggregate success rates do not reveal how a model moves. Table~\ref{tab:per_step} reports per-step statistics for drone action prediction, averaged over all predicted action sequences: the path ratio (predicted trajectory length divided by the ground-truth route length), the mean absolute lateral displacement $|\Delta x|$, the mean forward displacement $\Delta z$, and the mean absolute yaw change $|\Delta\theta|$.

\input{Tables/per_step_action}

Two models with similar success rates can behave very differently. Qwen2.5-VL-3B-Instruct and Llama-3.2-11B-Vision-Instruct both perform poorly ($\mathrm{SR}_{\mathrm{act}}$ of 0.133 and 0.247), but their profiles diverge sharply. Qwen2.5-VL-3B-Instruct produces unstable trajectories: its mean absolute yaw change reaches $20.31^\circ$ per step and its lateral displacement is the largest among all models, yielding a path ratio of 0.77, so the predicted route wanders even though its total length is below that of the reference. Llama-3.2-11B-Vision-Instruct instead produces overly conservative motion, with small lateral and angular changes and insufficient forward progress, which shows up as the lowest path ratio (0.59). The models with the highest success also have the most stable motion profiles: InternVL3.5-8B and Qwen3-VL-8B-Instruct combine near-unit path ratios with small yaw changes.

\subsection{Rollout-Level Failure Decomposition}

To characterise \emph{how} rollouts fail rather than only how often, we partition unsuccessful drone rollouts into three mutually exclusive categories. A rollout succeeds if its final planar position lies within $0.30$~m of the goal and it remains at least $0.30$~m from the obstacle throughout. Failures are \emph{action generation} failures when the rollout stays collision-free but does not reach the goal, \emph{collision} failures when it reaches the goal but violates the clearance constraint, and \emph{dual} failures when it satisfies neither condition.

\input{Tables/rollout_decomposition}

Table~\ref{tab:rollout_decomposition} shows that failure composition tracks model capability. For the weakest model, Qwen2.5-VL-3B-Instruct, $97.2\%$ of failures are action-generation failures and only $0.1\%$ are collisions, which is the direct signature of the safe-but-inactive behaviour noted in Sec.~\ref{sec4.2}: the model reliably avoids the obstacle and reliably fails to arrive. Models with higher success reduce action-generation failures without a proportional increase in collisions; Qwen3-VL-8B-Instruct and Idefics3-8B-Llama3 instead concentrate more failures in the dual category ($24.8\%$ and $31.0\%$), indicating that their remaining errors involve both planning and clearance.

We stress that these three categories describe rollout \emph{outcomes} and are not a causal attribution. Action-generation failures mix incorrect planning with execution-level effects, and the collision and dual categories provide only an observable proxy for rollout-related failures; separating planning from execution would require per-step comparison against the reference trajectory under identical rollout conditions, which we leave to future benchmark refinement.

\subsection{Cross-User Prediction Consistency}

Because VLAN presents every user profile with the identical visual scene and obstacle layout, any divergence in a model's accessibility prediction across profiles is attributable to user conditioning rather than to input differences. We exploit this to measure conditioning directly. Table~\ref{tab:cross_user} reports the joint distribution of the humanoid and BVI predictions on the same trajectory, where $P^{+}$ and $P^{-}$ denote a correct and an incorrect accessibility label respectively.

\input{Tables/cross_user_consistency}

The $P_{\mathrm{BVI}}^{-}P_{\mathrm{humanoid}}^{+}$ cell is the direct measure of BVI-specific conditioning failure: the model labels the trajectory correctly for the humanoid user but incorrectly for the BVI user on the same scene. Qwen3-VL-8B-Instruct shows the strongest conditioning, with only $2.2\%$ such failures, followed by InternVL3.5-8B ($7.6\%$) and Molmo2-8B ($10.1\%$). Idefics3-8B-Llama3 reaches $38.4\%$, indicating that its predictions change substantially across user profiles even when the visual input is held fixed. The near-zero $P_{\mathrm{BVI}}^{+}P_{\mathrm{humanoid}}^{-}$ column shows that the opposite failure, a correct BVI label paired with an incorrect humanoid label, is rare across all models. This asymmetry is itself informative: models err by failing to apply BVI-specific constraints rather than by inventing them where they do not apply.

\subsection{Closed-Source Reference Models}\label{app:closed_source}

The main results use open-weight models so that every number can be reproduced locally without external services. To provide commercial reference points, we additionally evaluate two closed-source models.

\input{Tables/closed_source}

As shown in Table~\ref{tab:closed_source}, the comparison does not yield a uniform ranking. Gemini 3.6 attains the best success rate on both action-prediction tasks and on the BVI setting, reaching $0.900$ on drone and $0.515$ on BVI. That advantage does not extend to every user, however: on the humanoid setting the best result belongs to an open-weight model, Llama-3.2-11B-Vision-Instruct at $0.565$, with Molmo2-8B second at $0.535$, and both commercial models fall below them. The wheelchair setting inverts the ordering again, where GPT-4o leads at $0.382$ and Gemini 3.6 drops to $0.205$. Closed-source capability is therefore not uniformly stronger across user profiles, which reinforces rather than contradicts the central observation of this paper: strong general capability does not transfer automatically to accessibility-conditioned navigation for every user type.

Because the subset is smaller than the full split, we quantify how closely it tracks the full-split conclusions. Across the seven open-weight models and five user settings, the maximum absolute deviation in success rate between the subset and the full split is $0.073$ (Idefics3-8B-Llama3 on the quadruped setting), with 34 of 35 cells below $0.05$. The top-4 model ordering is preserved in all five settings, and no model moves by more than one rank. The subset therefore preserves the comparative conclusions of the main text while keeping the commercial evaluation affordable.

\section{Annotation Procedure and Label Quality}
\label{app:annotation}

\subsection{Annotation Workflow}

The accessibility labels were assigned through a two-stage process designed to keep the criteria consistent at scale. During scene editing in Unity, each obstacle was labelled with user-specific accessibility labels for all five user types on every trajectory, based on the obstacle's physical properties (dimensions, clearance, height) and on each user's mobility constraints, including wheelchair turning radius, quadruped step height, and drone flight ceiling. After all trajectories had been generated, a front-end visualization tool was used to review every trajectory. This tool displays the scene, the obstacle layout, the trajectory path, and the assigned labels side by side, which allowed annotators to filter out low-quality data and to verify label consistency across the full corpus.

Because the criteria are geometric and physical rather than dependent on personal preference, annotation does not require lived experience of a specific disability: whether a wheelchair can physically pass through a gap of a given width, or whether an overhead branch obstructs a drone's flight path, follows from the user's body parameters and the obstacle geometry. We nonetheless note that involving participants with lived experience would strengthen the boundary cases between \textit{HighRisk} and \textit{Blocked}, and we identify this as a direction for a future iteration of the benchmark. The annotation pipeline could also be scaled by using large models to pre-label accessibility from obstacle geometry and user constraints, followed by human verification.

\subsection{Obstacle Recognition under the Rendering Gap}

Virtual obstacles inserted during Unity editing do not match the photometric realism of the Gaussian-splat reconstruction, which raises the question of whether models can still perceive them. We examined model outputs to test whether this rendering gap affects object understanding, and found that it does not appear to be the bottleneck.

Models name the obstacle they are reasoning about. Across sampled outputs, three models we inspected (Qwen2.5-VL-3B-Instruct, InternVL3.5-8B, and Llama-3.2-11B-Vision-Instruct) consistently produce the correct object category, for example \emph{``Walk forward about 0.5 meters. As you approach the traffic cone, keep slightly to the left to go around it.''} Quantitatively, models mention the obstacle in $70$--$96\%$ of pedestrian trajectories and achieve $61$--$75\%$ fine-grained semantic match across the sixteen obstacle categories.

That a model perceives the obstacle correctly does not mean it reasons about the obstacle correctly for a given user. The gap between these two abilities is visible in Table~\ref{tab:perception_action} of the main text, where $P^{+}A^{-}$ cases isolate trajectories on which the obstacle is identified correctly and the predicted action still fails. This separation is the evidence that obstacle perception is not the limiting factor for the models we evaluate.

\section{Dataset File Structure and Schema}
\label{app:schema}

The released dataset is organized as a directory tree rooted at the test split. Each scene directory contains a \texttt{KeyFrames/} subtree holding one subdirectory per trajectory, and a \texttt{Trajectory/} subtree holding one JSON file per trajectory. Each trajectory subdirectory under \texttt{KeyFrames/} contains the sampled keyframe metadata, the rendered RGB frames, and the accessibility annotations for that trajectory. This section documents the three JSON schemas that a user of the benchmark needs in order to construct targets, prompts, and evaluation references.

\subsection{File Schemas}

Table~\ref{tab:metadata_json_schema} documents the keyframe metadata. It records the scene and trajectory identifiers, the scene-level taxonomy fields used in Table~\ref{tab:scene_taxonomy_summary}, and the per-frame entries with world positions, orientations, and paths to the rendered RGB images.

\begin{table}[ht]
\centering
\caption{Schema of \texttt{metadata.json}.}
\label{tab:metadata_json_schema}
\small
\setlength{\tabcolsep}{4pt}
\renewcommand{\arraystretch}{0.96}
\begin{tabular}{p{0.28\linewidth} p{0.71\linewidth}}
\toprule
\textbf{Field} & \textbf{Description} \\
\midrule
\texttt{scene\_name} & Scene identifier. \\
\texttt{traj\_name} & Trajectory identifier within the scene. \\
\texttt{scene\_taxonomy} & Scene-level taxonomy fields including \texttt{lighting}, \texttt{scene\_type}, \texttt{path\_geometry}, and \texttt{weather}. \\
\texttt{keyframes} & Metadata for sampled keyframes, including selected frame indices and per-frame entries. \\
\midrule
\texttt{keyframes.frames[*]} & Each frame entry contains \texttt{order}, \texttt{frame\_idx}, \texttt{img\_name}, \texttt{position\_world}, \texttt{rotation\_world}, \texttt{rgb\_path}, and \texttt{src\_file}. \\
\bottomrule
\end{tabular}
\end{table}

Table~\ref{tab:trajectory_json_schema} documents the raw trajectory record. It lists the obstacles present in the scene along with their placement, the avoidance configuration used during generation, and the per-user accessibility annotations as originally captured in Unity.

\begin{table}[ht]
\centering
\caption{Schema of \texttt{trajectory.json}.}
\label{tab:trajectory_json_schema}
\small
\setlength{\tabcolsep}{4pt}
\renewcommand{\arraystretch}{0.96}
\begin{tabular}{p{0.2\linewidth} p{0.79\linewidth}}
\toprule
\textbf{Field} & \textbf{Description} \\
\midrule
\texttt{sceneName} & Scene identifier. \\
\texttt{obstacles} & List of scene obstacles used in the trajectory. \\
\texttt{trajectory} & Sequence of trajectory states. \\
\midrule
\texttt{obstacles[*]} & Each obstacle entry contains \texttt{objectId}, \texttt{displayName}, \texttt{position}, \texttt{rotation}, \texttt{scale}, \texttt{isGameObjectActive}, \texttt{participateInAvoidance}, \texttt{avoidanceMode}, \texttt{avoidanceOffset}, \texttt{hideDuringRecording}, \texttt{anchorType}, and \texttt{userAccessibility}. \\
\texttt{trajectory[*]} & Each trajectory point contains \texttt{id}, \texttt{img\_name}, \texttt{position}, and \texttt{rotation}. \\
\bottomrule
\end{tabular}
\end{table}

Table~\ref{tab:target_json_schema} documents the canonical target, which is the representation consumed by the benchmark. It is derived from the two preceding files and stores the primary obstacle, the representative frames used to ground accessibility reasoning, the goal, the base action decomposition, the final user-conditioned accessibility labels, and the expected output for each user type.

\begin{table*}[ht]
\centering
\caption{Schema of \texttt{target.json}.}
\label{tab:target_json_schema}
\small
\setlength{\tabcolsep}{4pt}
\renewcommand{\arraystretch}{0.96}
\begin{adjustbox}{width=\linewidth}
\begin{tabular}{p{0.32\linewidth} p{0.68\linewidth}}
\toprule
\textbf{Field} & \textbf{Description} \\
\midrule
\texttt{scene\_name} & Scene identifier. \\
\texttt{traj\_name} & Trajectory identifier. \\
\texttt{obstacle} & Obstacle information including \texttt{object\_id}, \texttt{display\_name}, \texttt{position\_world}, \texttt{avoidance\_mode}, \texttt{participate\_in\_avoidance}, and \texttt{semantic\_subtype}. \\
\texttt{representative\_frames} & Representative obstacle-related frames. Each entry includes \texttt{depth\_score}, \texttt{img\_name}, \texttt{obstacle\_local}, \texttt{order}, and \texttt{rgb\_path}. \\
\texttt{goal} & Goal metadata including \texttt{frame\_order} and \texttt{position\_world}. \\
\texttt{base\_actions} & Base trajectory action decomposition, including \texttt{observed\_prefix}, \texttt{full\_actions}, and \texttt{future\_reference}. \\
\texttt{corrected\_accessibility} & Final user-conditioned accessibility labels for \texttt{humanoid}, \texttt{blind}, \texttt{wheelchair}, \texttt{drone}, and \texttt{quadrupedRobot}. \\
\texttt{users} & User-specific benchmark targets. Language-guided users store \texttt{accessibility\_label}, \texttt{pass\_side}, and \texttt{ground\_truth\_instruction}. Action-prediction users store \texttt{accessibility\_label}, \texttt{observed\_prefix\_actions}, \texttt{ground\_truth\_actions}, and \texttt{rollout\_start\_pose\_world}; drone targets may additionally include \texttt{generation\_method}. \\
\bottomrule
\end{tabular}
\end{adjustbox}
\end{table*}

\subsection{Pre-Inference Data Processing Pipeline}
\label{app:pre_inference_pipeline}

Before model inference, the released dataset is transformed into two intermediate representations for each scene--trajectory pair: a \emph{canonical target} and a \emph{user-conditioned context}. This separation keeps the raw released data independent of any particular prompting format.

Algorithm~\ref{alg:build_targets} summarizes the target-construction stage. The pipeline reads the trajectory metadata, obstacle metadata, keyframe sequence, and user-conditioned accessibility annotations, and converts them into a unified benchmark target. The resulting target stores the obstacle identity, representative frames, goal metadata, base motion information, corrected accessibility labels, and the expected output for each user type. For language-guided users this output is a user-specific navigation instruction, while for action-prediction users it is a future action sequence.

Algorithm~\ref{alg:build_contexts} summarizes the context-construction stage. Given a canonical target, the pipeline derives the structured input context that is later rendered into the prompt for the multimodal model. This context includes the first-person image triplet, relative goal and obstacle geometry, prefix motion summaries, and user-specific navigation semantics. The humanoid context emphasizes metric route guidance, the blind context emphasizes step counts and clock directions, and the wheelchair context emphasizes route feasibility and clearance. For action-prediction users such as drones and quadruped robots, the context additionally includes rollout poses, local goal coordinates, obstacle-relative geometry, and pass-side constraints.

Together, the two stages decouple the released data from model prompting. The canonical target provides a stable task definition and evaluation reference, while the user-conditioned context provides a structured model input representation. This design makes the benchmark easier to maintain, extend, and reproduce across heterogeneous user types and model families.

\begin{algorithm}[t]
\caption{Build Canonical Targets Before Inference}
\label{alg:build_targets}
\begin{algorithmic}[1]
\Require Test split directory $\mathcal{D}$
\Require User set $\mathcal{U}=\{$humanoid, blind, wheelchair, drone, quadrupedRobot$\}$
\Ensure Canonical target file $T_{s,\tau}$ for each scene $s$ and trajectory $\tau$

\ForAll{scene directories $s \in \mathcal{D}$}
    \ForAll{trajectory directories $\tau$ in scene $s$}
        \State Load trajectory metadata $M_{s,\tau}$ from \texttt{metadata.json}
        \State Load trajectory state $R_{s,\tau}$ from \texttt{trajectory.json}
        \If{\texttt{target.json} already exists}
            \State Load released target $T_{s,\tau}$ directly
            \State Save $T_{s,\tau}$ to the artifacts directory
            \State \textbf{continue}
        \EndIf

        \State Extract keyframes $\{f_0,\dots,f_7\}$ from $M_{s,\tau}$
        \State Extract obstacle set $\mathcal{O}_{s,\tau}$ from $R_{s,\tau}$
        \State Select the primary obstacle $o^\star$ used for accessibility reasoning
        \State Infer semantic subtype of $o^\star$
        \State Compute representative frames around $o^\star$
        \State Compute action sequence $A_{s,\tau}$ from keyframe poses
        \State Split $A_{s,\tau}$ into:
        \Statex \hspace{1.5em} observed prefix actions $A^{\mathrm{obs}}_{s,\tau}$
        \Statex \hspace{1.5em} future reference actions $A^{\mathrm{ref}}_{s,\tau}$

        \State Read per-user accessibility labels from trajectory annotations
        \State Normalize user keys to the public schema:
        \Statex \hspace{1.5em} \texttt{humanoid}, \texttt{blind}, \texttt{wheelchair}, \texttt{drone}, \texttt{quadrupedRobot}

        \ForAll{$u \in \mathcal{U}$}
            \If{$u=$ humanoid}
                \State Generate humanoid navigation target instruction
            \ElsIf{$u=$ blind}
                \State Derive blind accessibility label and generate BVI-style instruction
            \ElsIf{$u=$ wheelchair}
                \State Derive wheelchair accessibility label and generate wheelchair-specific instruction
            \ElsIf{$u=$ quadrupedRobot}
                \State Use $A^{\mathrm{ref}}_{s,\tau}$ as future action target
            \ElsIf{$u=$ drone}
                \State Generate drone future actions according to accessibility mode
                \Statex \hspace{1.5em} (e.g., passable straight-line, overhead bypass, or stop)
            \EndIf
        \EndFor

        \State Pack scene ID, trajectory ID, obstacle metadata, representative frames,
        \Statex \hspace{1.5em} goal metadata, base actions, corrected accessibility, and all user targets
        \State Save the canonical target file $T_{s,\tau}$
    \EndFor
\EndFor
\end{algorithmic}
\end{algorithm}

\begin{algorithm}[t]
\caption{Build User-Conditioned Contexts Before Inference}
\label{alg:build_contexts}
\begin{algorithmic}[1]
\Require Test split directory $\mathcal{D}$
\Require Canonical targets $\{T_{s,\tau}\}$
\Ensure Context file $C_{s,\tau}$ for each scene $s$ and trajectory $\tau$

\ForAll{scene directories $s \in \mathcal{D}$}
    \ForAll{trajectory directories $\tau$ in scene $s$}
        \State Load canonical target $T_{s,\tau}$
        \State Load keyframe metadata $M_{s,\tau}$ from \texttt{metadata.json}
        \State Extract input images $\{I_1,I_2,I_3\}$ from the first three keyframes
        \State Extract rollout start frame and goal frame from the keyframe sequence
        \State Read primary obstacle metadata from $T_{s,\tau}$

        \State Compute shared geometric quantities:
        \Statex \hspace{1.5em} goal relative direction and distance
        \Statex \hspace{1.5em} obstacle relative direction and distance
        \Statex \hspace{1.5em} safer bypass side / clock-side hints

        \State Build humanoid context:
        \Statex \hspace{1.5em} accessibility label
        \Statex \hspace{1.5em} prefix motion summary
        \Statex \hspace{1.5em} goal summary
        \Statex \hspace{1.5em} obstacle summary

        \State Build blind context:
        \Statex \hspace{1.5em} step-based goal distance
        \Statex \hspace{1.5em} clock-direction goal and obstacle hints
        \Statex \hspace{1.5em} blind-specific risk and pass-side hints

        \State Build wheelchair context:
        \Statex \hspace{1.5em} route accessibility summary
        \Statex \hspace{1.5em} clearance / feasibility summary
        \Statex \hspace{1.5em} wheelchair-specific risk hints

        \ForAll{$u \in \{$drone, quadrupedRobot$\}$}
            \State Build action-planning context for $u$:
            \Statex \hspace{1.5em} observed prefix actions
            \Statex \hspace{1.5em} rollout start pose
            \Statex \hspace{1.5em} goal position in world and local coordinates
            \Statex \hspace{1.5em} obstacle position in world and local coordinates
            \Statex \hspace{1.5em} required pass side
            \Statex \hspace{1.5em} bypass corridor and directional hints
        \EndFor

        \State Pack:
        \Statex \hspace{1.5em} scene ID, trajectory ID, input image list, obstacle metadata,
        \Statex \hspace{1.5em} corrected accessibility labels, and all user-conditioned contexts
        \State Save context file $C_{s,\tau}$
    \EndFor
\EndFor
\end{algorithmic}
\end{algorithm}                                                                                                                                                                                                                                                                
                                                                                                                                                                                                                                                                                                                       
  Before model inference, the released dataset is transformed into two intermediate representations: a \emph{canonical target} and a \emph{user-conditioned context} for each scene--trajectory pair.                                                                                                                  
  Algorithm~\ref{alg:build_targets} summarizes the target-construction stage.                                                                                                                                                                                                                                          
  At this stage, the pipeline reads the trajectory metadata, obstacle metadata, keyframe sequence, and user-conditioned accessibility annotations, and converts them into a unified benchmark target.                                                                                                                  
  The resulting target representation stores the obstacle identity, representative frames, goal metadata, base motion information, corrected accessibility labels, and the expected output for each user type.                                                                                                         
  For language-guided users, this output is a user-specific navigation instruction, while for action-prediction users it is a future action sequence.

  Algorithm~\ref{alg:build_contexts} summarizes the context-construction stage.
  Given a canonical target, the pipeline derives the structured input context that will later be rendered into prompts for the multimodal model.
  This context includes the first-person image triplet, relative goal and obstacle geometry, prefix motion summaries, and user-specific navigation semantics.
  For example, the humanoid context emphasizes metric route guidance, the blind context emphasizes step counts and clock directions, and the wheelchair context emphasizes route feasibility and clearance.
  For action-prediction users such as drones and quadruped robots, the context further includes rollout poses, local goal coordinates, obstacle-relative geometry, and pass-side constraints.

\clearpage
\flushbottom

%% file: Tables/drone_differential.tex
\begin{table*}[ht]
\centering
\caption{\textbf{Differential analysis of embodiment-specific difficulty} on the full test split and on the 1,371 unconstrained trajectories (those where the obstacle is irrelevant to the drone at its flight altitude). $\Delta$ denotes the change in $\mathrm{SR}_{\mathrm{act}}$ when the 560 constrained trajectories are removed, and $\Delta\Delta = \Delta_{\mathrm{drone}} - \Delta_{\mathrm{quadruped}}$. A positive $\Delta\Delta$ indicates that removing constrained trajectories benefits the drone more than the quadruped robot.}
\label{tab:drone_differential}
\small
\setlength{\tabcolsep}{6pt}
\renewcommand{\arraystretch}{0.96}
\begin{adjustbox}{width=\linewidth}
\begin{tabular}{l c c c c c c c}
\toprule
 & \multicolumn{3}{c}{\textbf{Drone}} & \multicolumn{3}{c}{\textbf{Quadruped Robot}} & \\
\cmidrule(lr){2-4} \cmidrule(lr){5-7}
\textbf{Model} & Full & Unconstrained & $\Delta$ & Full & Unconstrained & $\Delta$ & $\Delta\Delta$ \\
\midrule
Qwen2.5-VL-3B-Instruct        & 0.133 & 0.167 & $+0.034$ & 0.053 & 0.052 & $-0.001$ & $+0.035$ \\
Qwen2.5-VL-7B-Instruct        & 0.571 & 0.624 & $+0.054$ & 0.384 & 0.373 & $-0.012$ & $+0.065$ \\
Qwen3-VL-8B-Instruct          & 0.622 & 0.656 & $+0.034$ & 0.474 & 0.470 & $-0.003$ & $+0.038$ \\
InternVL3.5-8B                & 0.642 & 0.729 & $+0.088$ & 0.393 & 0.403 & $+0.010$ & $+0.078$ \\
Idefics3-8B-Llama3            & 0.554 & 0.654 & $+0.100$ & 0.308 & 0.309 & $+0.002$ & $+0.098$ \\
Molmo2-8B                     & 0.289 & 0.363 & $+0.074$ & 0.147 & 0.136 & $-0.011$ & $+0.085$ \\
Llama-3.2-11B-Vision-Instruct & 0.247 & 0.317 & $+0.071$ & 0.048 & 0.045 & $-0.003$ & $+0.074$ \\
\bottomrule
\end{tabular}
\end{adjustbox}
\end{table*}

%% file: Tables/results_by_lighting.tex
  \begin{table*}[ht]
  \centering
  \caption{\textbf{Results by lighting condition.}
  We report Accessible Success Rate (AccSR) under four lighting conditions.
  }
  \label{tab:results_by_lighting}
  \small
  \setlength{\tabcolsep}{4pt}
  \renewcommand{\arraystretch}{0.96}
  \begin{adjustbox}{width=\linewidth}
  \begin{tabular}{p{0.36\linewidth} c c c c c}
  \toprule
  \textbf{Model} & \textbf{Params} & \textbf{Daytime} $\uparrow$ & \textbf{Nighttime} $\uparrow$ & \textbf{Low Vision} $\uparrow$ & \textbf{Dusk} $\uparrow$ \\
  \midrule
  Qwen2.5-VL-3B-Instruct & 3B & 0.159 & 0.148 & 0.127 & 0.047 \\
  Qwen2.5-VL-7B-Instruct & 7B & 0.335 & 0.347 & 0.259 & 0.299 \\
  Qwen3-VL-8B-Instruct & 8B & \underline{0.390} & \textbf{0.423} & \underline{0.390} & \underline{0.327} \\
  InternVL3.5-8B & 8B & \textbf{0.433} & \underline{0.420} & \textbf{0.412} & \textbf{0.368} \\
  Idefics3-8B-Llama3 & 8B & 0.260 & 0.272 & 0.232 & 0.233 \\
  Molmo2-8B & 8B & 0.275 & 0.217 & 0.220 & 0.195 \\
  Llama-3.2-11B-Vision-Instruct & 11B & 0.190 & 0.209 & 0.195 & 0.127 \\
  \bottomrule
  \end{tabular}
  \end{adjustbox}
  \end{table*}

%% file: Tables/results_by_geo.tex
  \begin{table*}[ht]
  \centering
  \caption{\textbf{Results by path geometry.}
  We report Accessible Success Rate (AccSR) under four path-geometry categories. Higher is better.
  }
  \label{tab:results_by_path_geometry}
  \small
  \setlength{\tabcolsep}{4pt}
  \renewcommand{\arraystretch}{0.96}
  \begin{adjustbox}{width=\linewidth}
  \begin{tabular}{p{0.36\linewidth} c c c c c}
  \toprule
  \textbf{Model} & \textbf{Params} & \textbf{Wide} $\uparrow$ & \textbf{Open} $\uparrow$ & \textbf{Narrow} $\uparrow$ & \textbf{Constrained} $\uparrow$ \\
  \midrule
  Qwen2.5-VL-3B-Instruct & 3B & 0.160 & 0.116 & 0.162 & 0.126 \\
  Qwen2.5-VL-7B-Instruct & 7B & 0.339 & 0.344 & 0.306 & 0.314 \\
  Qwen3-VL-8B-Instruct & 8B & \underline{0.412} & \underline{0.397} & \underline{0.367} & \underline{0.358} \\
  InternVL3.5-8B & 8B & \textbf{0.432} & \textbf{0.419} & \textbf{0.425} & \textbf{0.418} \\
  Idefics3-8B-Llama3 & 8B & 0.275 & 0.273 & 0.235 & 0.225 \\
  Molmo2-8B & 8B & 0.262 & 0.230 & 0.275 & 0.245 \\
  Llama-3.2-11B-Vision-Instruct & 11B & 0.209 & 0.193 & 0.160 & 0.180 \\
  \bottomrule
  \end{tabular}
  \end{adjustbox}
  \end{table*}

%% file: Tables/per_step_action.tex
\begin{table*}[ht]
\centering
\caption{\textbf{Per-step action statistics for drone action prediction.} Statistics are averaged over all predicted action sequences. Path ratio is the predicted trajectory length divided by the ground-truth route length. $|\Delta x|$ and $\Delta z$ are in meters, $|\Delta\theta|$ in degrees.}
\label{tab:per_step}
\small
\setlength{\tabcolsep}{10pt}
\renewcommand{\arraystretch}{0.96}
\begin{adjustbox}{width=\linewidth}
\begin{tabular}{l c c c c c}
\toprule
\textbf{Model} & $\mathbf{SR}_{\mathbf{act}}\uparrow$ & \textbf{Path ratio} & $\mathbf{mean}\,|\Delta x|$ & $\mathbf{mean}\,\Delta z$ & $\mathbf{mean}\,|\Delta\theta|$ \\
\midrule
Qwen2.5-VL-3B-Instruct        & 0.133 & 0.77 & 0.196 & 0.170 & 20.31 \\
Qwen2.5-VL-7B-Instruct        & 0.571 & 0.96 & 0.081 & 0.385 & 0.19 \\
Qwen3-VL-8B-Instruct          & 0.622 & 1.01 & 0.116 & 0.396 & 0.58 \\
InternVL3.5-8B                & 0.642 & 1.04 & 0.132 & 0.402 & 1.21 \\
Idefics3-8B-Llama3            & 0.554 & 1.05 & 0.114 & 0.399 & 0.01 \\
Molmo2-8B                     & 0.289 & 1.07 & 0.167 & 0.395 & 0.58 \\
Llama-3.2-11B-Vision-Instruct & 0.247 & 0.59 & 0.056 & 0.228 & 0.28 \\
\bottomrule
\end{tabular}
\end{adjustbox}
\end{table*}

%% file: Tables/rollout_decomposition.tex
\begin{table*}[ht]
\centering
\caption{\textbf{Rollout-level decomposition of drone failures} (\%). Successful rollouts require a final position within $0.30$~m of the goal and at least $0.30$~m clearance from the obstacle throughout. \emph{Action gen}: collision-free but does not reach the goal. \emph{Collision}: reaches the goal but violates clearance. \emph{Dual}: satisfies neither condition. The three categories are operational rollout outcomes, not a causal attribution between planning and execution.}
\label{tab:rollout_decomposition}
\small
\setlength{\tabcolsep}{10pt}
\renewcommand{\arraystretch}{0.96}
\begin{adjustbox}{width=\linewidth}
\begin{tabular}{l c c c c}
\toprule
\textbf{Model} & $\mathbf{SR}_{\mathbf{act}}$ & \textbf{Action gen.} & \textbf{Collision} & \textbf{Dual} \\
\midrule
Qwen2.5-VL-3B-Instruct        & 0.133 & 97.2 & 0.1  & 2.6  \\
Qwen2.5-VL-7B-Instruct        & 0.571 & 72.3 & 10.4 & 17.4 \\
Qwen3-VL-8B-Instruct          & 0.622 & 67.4 & 7.8  & 24.8 \\
InternVL3.5-8B                & 0.642 & 79.0 & 9.4  & 11.6 \\
Idefics3-8B-Llama3            & 0.554 & 60.9 & 8.1  & 31.0 \\
Molmo2-8B                     & 0.289 & 87.7 & 3.6  & 8.7  \\
Llama-3.2-11B-Vision-Instruct & 0.247 & 91.5 & 2.5  & 6.0  \\
\bottomrule
\end{tabular}
\end{adjustbox}
\end{table*}

%% file: Tables/cross_user_consistency.tex
\begin{table*}[ht]
\centering
\caption{\textbf{Cross-analysis of accessibility prediction consistency across user profiles.} Each cell is the joint probability of the humanoid and BVI predictions on the same trajectory, where $P^{+}$ and $P^{-}$ denote a correct and an incorrect accessibility label. $P_{\mathrm{BVI}}^{-}P_{\mathrm{humanoid}}^{+}$ measures BVI-specific user-conditioning failure.}
\label{tab:cross_user}
\small
\setlength{\tabcolsep}{10pt}
\renewcommand{\arraystretch}{0.96}
\begin{adjustbox}{width=\linewidth}
\begin{tabular}{l c c c c}
\toprule
\textbf{Model} & $\mathbf{P^{+}_{\mathrm{BVI}}P^{+}_{\mathrm{humanoid}}}$ & $\mathbf{P^{-}_{\mathrm{BVI}}P^{+}_{\mathrm{humanoid}}}$ & $\mathbf{P^{+}_{\mathrm{BVI}}P^{-}_{\mathrm{humanoid}}}$ & $\mathbf{P^{-}_{\mathrm{BVI}}P^{-}_{\mathrm{humanoid}}}$ \\
\midrule
Qwen2.5-VL-3B-Instruct & 0.692 & 0.281 & 0.010 & 0.017 \\
Qwen2.5-VL-7B-Instruct & 0.692 & 0.293 & 0.008 & 0.007 \\
Qwen3-VL-8B-Instruct   & 0.964 & 0.022 & 0.007 & 0.007 \\
InternVL3.5-8B         & 0.908 & 0.076 & 0.007 & 0.008 \\
Idefics3-8B-Llama3     & 0.600 & 0.384 & 0.008 & 0.008 \\
Molmo2-8B              & 0.885 & 0.101 & 0.007 & 0.008 \\
\bottomrule
\end{tabular}
\end{adjustbox}
\end{table*}

%% file: Tables/closed_source.tex
\begin{table*}[ht]
\centering
\caption{\textbf{All evaluated models on the fixed 200-trajectory subset.} All nine models are evaluated under one identical protocol on a 200-trajectory subset of the VLAN test split (two trajectories per scene, seed 42), so that every entry in this table shares the same evaluation budget. These numbers are \emph{not} directly comparable with Tables~\ref{tab:simplified_action_results}--\ref{tab:accessible_crossuser}, which use the full 1,931-trajectory split. \textbf{Bold} marks the best and \underline{underline} the second-best value in each column; ties are marked for all tied models.}
\label{tab:closed_source}
\small
\setlength{\tabcolsep}{9pt}
\renewcommand{\arraystretch}{1.0}
\begin{adjustbox}{width=\linewidth}
\begin{tabular}{l c c c c c}
\toprule
\textbf{Model} & \textbf{Drone} & \textbf{Quadruped} & \textbf{Humanoid} & \textbf{BVI} & \textbf{Wheelchair} \\
\midrule
Qwen2.5-VL-3B-Instruct        & 0.125 & 0.045 & 0.525 & 0.240 & \underline{0.335} \\
Qwen2.5-VL-7B-Instruct        & 0.600 & 0.425 & 0.270 & 0.255 & 0.190 \\
Qwen3-VL-8B-Instruct          & 0.620 & \underline{0.465} & 0.410 & \underline{0.425} & 0.115 \\
InternVL3.5-8B                & \underline{0.685} & 0.380 & 0.390 & 0.355 & 0.330 \\
Idefics3-8B-Llama3            & 0.540 & 0.235 & 0.445 & 0.115 & 0.005 \\
Molmo2-8B                     & 0.250 & 0.175 & \underline{0.535} & 0.130 & 0.280 \\
Llama-3.2-11B-Vision-Instruct & 0.205 & 0.040 & \textbf{0.565} & 0.175 & 0.095 \\
GPT-4o                        & 0.565 & 0.415 & 0.465 & \underline{0.425} & \textbf{0.382} \\
Gemini 3.6                    & \textbf{0.900} & \textbf{0.546} & 0.470 & \textbf{0.515} & 0.205 \\
\bottomrule
\end{tabular}
\end{adjustbox}
\end{table*}